\pdfoutput=1

\documentclass[11pt]{article}

\newcommand{\revoption}{final}
\usepackage[\revoption]{acl}

\usepackage{times}
\usepackage{latexsym}

\usepackage[T1]{fontenc}

\usepackage[utf8]{inputenc}

\usepackage{microtype}

\usepackage{adjustbox}
\usepackage{array}

\newcolumntype{R}[2]{%
    >{\adjustbox{angle=#1,lap=\width-(#2)}\bgroup}%
    l%
    <{\egroup}%
}
\newcommand*\rot{\multicolumn{1}{R{35}{1em}}}

\usepackage{adjustbox}
\usepackage{graphicx}
\usepackage{multirow}
\usepackage{booktabs}
\usepackage{todonotes}
\usepackage{float}
\usepackage{xstring}
\usepackage{enumitem}
\usepackage[bottom]{footmisc}

\definecolor{lightblue}{HTML}{E0ECF7}
\definecolor{darkblue}{HTML}{092E6B}
\newcommand{\win}[1]{{\colorbox{lightblue}{\sf #1}}}
\newcommand{\lose}[1]{{\colorbox{darkblue}{\sf \color{white}{#1}}}}

\usepackage{soul}
\usepackage{etoolbox}
\makeatletter
\patchcmd{\SOUL@ulunderline}{\dimen@}{\SOUL@dimen}{}{}
\patchcmd{\SOUL@ulunderline}{\dimen@}{\SOUL@dimen}{}{}
\patchcmd{\SOUL@ulunderline}{\dimen@}{\SOUL@dimen}{}{}
\newdimen\SOUL@dimen
\makeatother

\title{Language Models that Seek for Knowledge:\\ Modular Search \& Generation for Dialogue and Prompt Completion}

\author{ Kurt Shuster\\
  Facebook AI Research\\
 \\\And
  Mojtaba Komeili\\
  Facebook AI Research\\
 \\\And
 Leonard Adolphs \thanks{\hspace{.5em}Work done during a Facebook AI Research internship.} \\
  ETH Zürich \\
 \\\AND
Stephen Roller\\
  Facebook AI Research\\
  \\\And
  Arthur Szlam\\
  Facebook AI Research\\
  \\\And
  Jason Weston \\
  Facebook AI Research\\
}

\begin{document}
\maketitle
\begin{abstract}
Language models (LMs) have recently been shown to generate more factual responses  
by employing  modularity \cite{zhou2021think} in combination with retrieval \cite{adolphs2021reason}. We extend the recent approach of \citet{adolphs2021reason} to include internet search as a module. Our SeeKeR 
(\underline{Se}arch-\underline{e}ngine$\rightarrow$\underline{K}nowledg\underline{e}$\rightarrow$\underline{R}esponse)
method thus applies a single LM to three modular tasks in succession:
search, generating knowledge, and generating a final response.
We show that, when
using SeeKeR as a dialogue model, it outperforms the state-of-the-art model BlenderBot 2 \cite{bb2}
on open-domain knowledge-grounded conversations for the same number of parameters, in terms of consistency, knowledge and per-turn engagingness. 
SeeKeR applied to topical prompt completions as a standard language model 
outperforms  GPT2 \cite{radford2019language} and 
GPT3 \cite{brown2020language}
in terms of factuality and topicality, despite GPT3 being a vastly larger model. Our code and models are made publicly available\footnote{\tiny\url{http://parl.ai/projects/seeker}}.
\end{abstract}

\section{Introduction}

Standard large language models are known to 
generate fluent but factually incorrect statements, a problem that is not solved by just increasing their size
\cite{hallucination_conversation}. Additionally, as their knowledge is frozen in time from the point when they were trained, they can never learn new facts -- the newest information they  have will be from the date that the training set was constructed.  
Several recent advances have tried to tackle aspects of these problems. Neural retrieval models have augmented seq2seq models with access to a large fixed corpus of knowledge \citep{lee-etal-2019-latent, rag_dpr}. However, aggregating information from multiple retrieved documents is a difficult problem \cite{izacard-grave-2021-leveraging} which may result in incorporating parts of multiple documents into one factually incorrect response. A modular approach which first finds the relevant parts of the documents and then generates the final response has been shown to help alleviate this problem \cite{adolphs2021reason}. However, none of those methods
can incorporate new information, which has been studied in separate work that augments generations with internet search \cite{DBLP:journals/corr/abs-2107-07566}.

\begin{figure}[t!]
    \centering
     \includegraphics[width=1.0\linewidth]{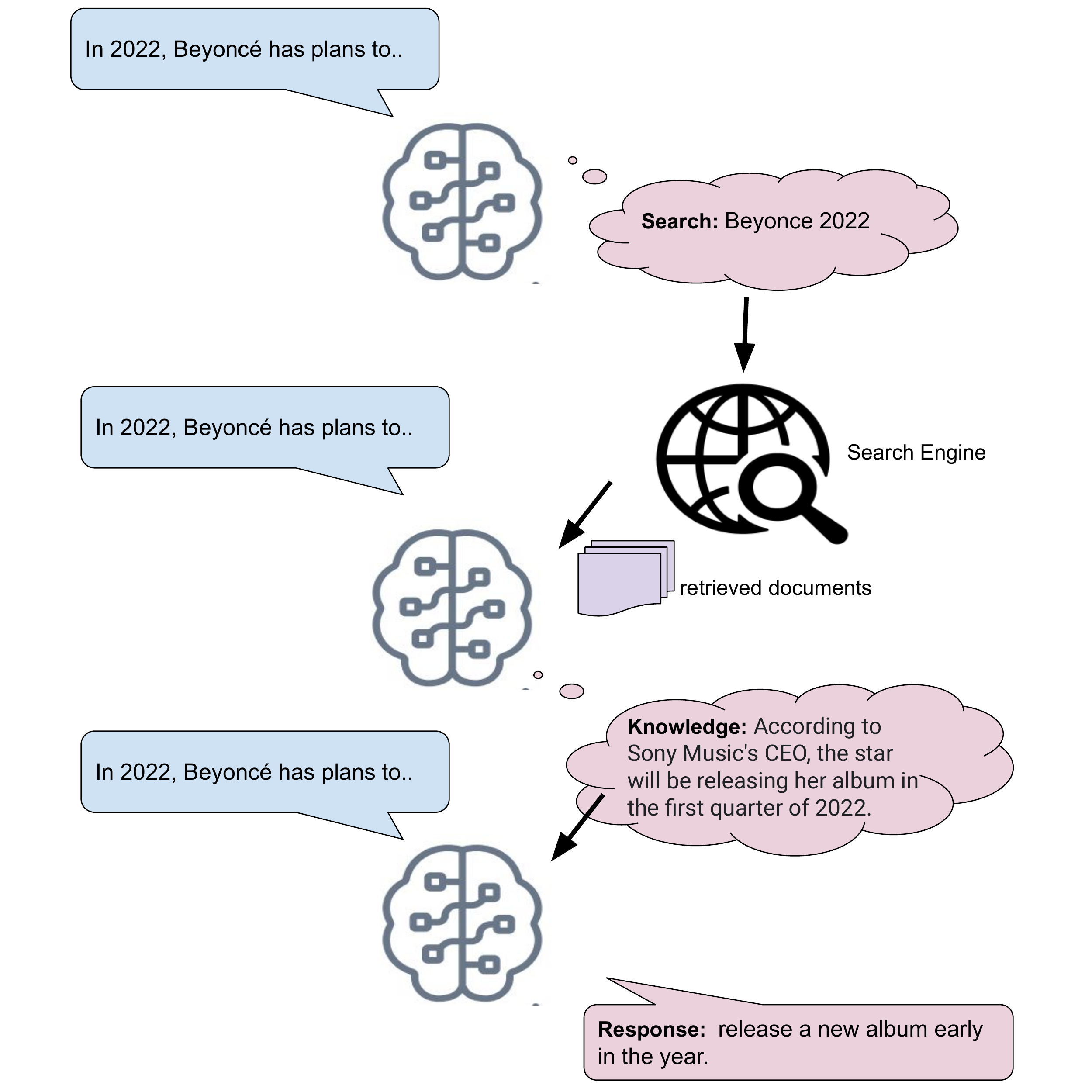}
     \caption{The modular \underline{Se}arch-\underline{e}ngine $\rightarrow$ \underline{K}nowledg\underline{e} $\rightarrow$ \underline{R}esponse (SeeKeR) Language Model. 
     A single transformer architecture is called successively to invoke three different modules: search, generate knowledge, and generate final response. The output of each module is input to the next, in addition to the original context.
     \label{fig:SeeKeR}
     }
\end{figure}

In this paper, we explore a modular architecture that tries to mix the best elements of these different existing solutions. A single transformer architecture is used iteratively to perform three modular tasks: search, generate knowledge, and generate a final response, where the output of each module is fed as additional input to the next, as in Figure \ref{fig:SeeKeR}. 
 The first step, given the input context, generates a relevant search query for an internet search engine, while the second step is fed the returned documents and generates their most relevant portion. The last step uses that knowledge to produce its final response.
 By decomposing this difficult problem into three manageable steps, pertinent up-to-date information can be incorporated into the final language model generation.

We apply our modular 
\underline{Se}arch-\underline{e}ngine$ \rightarrow$ \underline{K}nowledg\underline{e} $\rightarrow$ \underline{R}esponse
(SeeKeR) language model to the tasks of dialogue and prompt completion, after pre-training and fine-tuning on a variety of knowledge-intensive datasets.
In open-domain dialogue, we show this approach outperforms
the state-of-the-art BlenderBot 2 model of 
\citet{bb2} 
according to human ratings of consistency, knowledge and 
per-turn engagingness.

We  test the ability of SeeKeR to perform general -- but up-to-date -- language modeling. To do this we construct topical prompts on subjects that were in the news in January 2022, which is data that the model itself has not been trained on. With SeeKeR's ability to incorporate information via web search, it outperforms GPT2 \cite{radford2019language} and 
GPT3 \cite{brown2020language} in terms of factuality and topicality according to human raters.

\section{Related Work}

Our work builds on the knowledge to response (K2R) technique \cite{adolphs2021reason} which decomposes a dialogue model into two stages: generating a knowledge sequence, followed by generating a response sequence, conditioned on the knowledge. This was applied successfully to Wizard of Wikipedia \cite{dinan2018wizard}, QA \cite{lee-etal-2019-latent} and LIGHT tasks \cite{urbanek2019learning}. We expand on this approach by adding the additional module of internet search and then applying that to full open-domain dialogue and general language modeling. 

In the dialogue space, the most natural comparison to our approach is BlenderBot 2 (BB2) \cite{bb2}. BB2 grounds on retrieval from the internet  for open-domain dialogue tasks \cite{DBLP:journals/corr/abs-2107-07566}, but does not use a modular approach to generate knowledge, instead applying
the fusion-in-decoder (FiD) method \cite{izacard2020distilling} to output a response directly given the retrieved documents. They, as well as others \cite{lee2022empirical}, report that their method can have the problems of 
either mixing up facts together incorrectly or generating a generic response that ignores the knowledge, which our method attempts to address.
Another recent approach that uses information retrieval is LaMDA \cite{thoppilan2022lamda}, where the retrieval engine returns pertinent information (rather than a set of documents) and is considered a separate black box. LaMDA is not openly available and cannot be compared to.
WebGPT \cite{nakano2021webgpt} also applies internet search to QA tasks, as does the work of \citet{lazaridou2022internetaugmented}; neither applies to dialogue or general LM tasks, and neither work is openly available.

In the language modeling space, there is a large body of work 
on nearest neighbor and cache-based language modeling \cite{Khandelwal2020Generalization,grave2016improving,merity2016pointer,kh2020nearest,yogatama2021adaptive}
for accessing a large
set of documents. Recently, RETRO \cite{borgeaud2021improving} used retrieval over a database of trillions of tokens.
Those works do not use internet search, but rather perform their own retrieval method via a transformer model together with  nearest neighbor lookup. As the database is fixed, that means it would not be up to date  with the latest knowledge and current events.
Some recent methods have also attempted to adapt knowledge through editing and tuning of language model variants \cite{de2021editing,mitchell2021fast}.

\section{SeeKeR Model}

The SeeKeR model we introduce in this paper has the architecture of a standard transformer \cite{NIPS2017_3f5ee243}, except that this same encoder-decoder (for dialogue) or decoder-only (for language modeling) model is used in a modular way multiple times. For each module, special tokens are used in the encoder (or decoder) to indicate which module is being invoked. The output of each module is input into the next, along with the original context.

SeeKeR consists of three modules, which are invoked sequentially:

\paragraph{Search Module} Given the encoded input context, a search query is generated. This is fed into a search engine, which returns results in the form of a set of documents.  Following \citet{DBLP:journals/corr/abs-2107-07566}, in our  experiments (unless stated otherwise) we employ the Bing Web Search API\footnote{\tiny{\url{https://www.microsoft.com/en-us/bing/apis/bing-web-search-api}}} to retrieve documents, and then filter that set of documents by intersecting with Common Crawl \cite{wenzek2019ccnet}, and keep the top 5.

\paragraph{Knowledge Module} Given the encoded input context, and a set of retrieved documents, a knowledge response is generated. This consists of one or more relevant phrases or sentences from the retrieved documents. For encoder-decoder models, the documents and context are encoded using the fusion-in-decoder (FiD) method \cite{izacard2020distilling}; for decoder-only models, we pack and prepend the documents to the input context.
Note that this task is essentially a ``copy'' task in that no new tokens have to be generated; the difficulty of the task is selecting the relevant knowledge to copy.

\paragraph{Response Module} Given the encoded input context concatenated with the knowledge response, the final response is generated.  The module must consider relevant context and knowledge while generating a new fluent continuation to the input. The extraction of relevant knowledge by the previous modules makes this task easier; in contrast, a conventional seq2seq model has to solve all these tasks (knowledge acquisition, synthesis, and final response generation) at once.

\subsection{Architecture and Pre-Training} \label{sec:pre-train}

For our standard language modeling experiments, we consider the GPT2 transformer  \cite{radford2019language} as a base model, and  fine-tune it to become a SeeKeR model (see \autoref{sec:lm_tasks}); we do not perform any pre-training of our own in this case. We can thus directly compare to GPT2, with the same model size and architecture. We consider medium, large and XL  (345M, 762M and 1.5B parameters) models in our experiments.

For our dialogue experiments, we employ a 2.7B parameter transformer encoder-decoder model.
To pre-train our model we consider combining two different pre-training datasets for language-modeling and for dialogue, using the training method of \citet{lewis2019bart}:

\paragraph{pushshift.io Reddit}
We use a variant of Reddit discussions, which has also been used in several existing studies, particularly for training BlenderBot 1 and 2 \cite{roller-etal-2021-recipes}. The setup requires training to generate a comment conditioned on the full thread leading up to the comment. Following \citet{humeau2019polyencoder}, this is a previously existing Reddit dataset extracted and obtained by a third party and made available on pushshift.io \citep{baumgartner2020pushshift}, spanning 1.5B training examples from Reddit obtained from PushShift\footnote{\tiny\url{https://files.pushshift.io/reddit/}} through July 2019. A number of heuristic rules have been used to filter and clean the dataset; see \citet{roller-etal-2021-recipes} for details.

\paragraph{RoBERTa+CC100en}
We use the same data used to train the BASE language model \citep{lewis2021base}, which consists of approximately 100B tokens, combining corpora used in RoBERTa \citep{liu2019roberta} with the English subset of the CC100 corpus \citep{conneau2019unsupervised}.

We compare pre-training only on dialogue modeling (pushshift.io Reddit, as in \cite{roller-etal-2021-recipes}) to pre-training on both language modeling and dialogue modeling tasks; we refer to the latter as R2C2 (pushshift.io \underline{R}eddit, \underline{R}oBERTa + \underline{CC}100en). 
Full details, including architectural and pre-training hyperparameters, are discussed in \autoref{sec:arch_pt_params_appendix}.

\subsection{SeeKeR Tasks for Dialogue} \label{sec:dialog_ft_tasks}

We consider a number of dialogue-based fine-tuning tasks to enable our model to perform well for each of the three modules. 

\paragraph{Search Module Tasks} 
We use data from the Wizard of Internet (WizInt) task \cite{DBLP:journals/corr/abs-2107-07566} which consists of 
8,614 training dialogues containing 42,306 human-authored relevant search queries given the dialogue contexts. We can use the search query data as targets  to directly train the search module in a supervised fashion. 
We append special tokens to the input context to indicate that the transformer is performing the search task, via predicting a relevant search query.

\paragraph{Knowledge Module Tasks}
We multi-task several knowledge-intensive NLP tasks, where the target for the model is the ``knowledge'' that will be used to generate the final response. We first employ 
knowledge grounded dialogue datasets that contain annotations of the gold knowledge used:  Wizard of Internet \cite{DBLP:journals/corr/abs-2107-07566}
and Wizard of Wikipedia (WoW) \cite{dinan2018wizard}.
We then use several QA tasks:  SQuAD \cite{rajpurkar2016squad}, TriviaQA \cite{joshi2017triviaqa},
Natural Questions (NQ) \cite{kwiatkowski2019natural},
and MS MARCO \cite{nguyen2016ms}.
We use the ``Natural Language Generation'' competition track (NLGen v2.1) of MS MARCO, in which the annotator must ``provide your answer in a way in which it could be read from a smart speaker and make sense without any additional context''\footnote{\tiny{\url{https://microsoft.github.io/msmarco/}}}. As such, the original targets do not have direct overlap with one of the input documents, so we modify the task to satisfy this constraint by finding the highest overlapping input sentence with the answer, and make that the target instead. If the F1 overlap is less than 0.5 we drop the example, leaving  281,658 examples out of the original 808,731.
For NQ, we use three different settings: with all documents as input, with only the gold document, and with a sampled dialogue history context, following \cite{adolphs2021reason}. Finally, we can employ conventional dialogue tasks in this setting as well -- PersonaChat \cite{zhang2018personalizing}, Empathetic Dialogues (ED) \cite{rashkin2019empathetic} and Blended Skill Talk (BST) \cite{smith2020bst} --
by using the same procedure as in \cite{adolphs2021reason}: we extract an entity from the original dialogue response that also appears in the context, and set that as the knowledge target for training.  We also employ the Multi-Session Chat (MSC) \cite{xu2021beyond} task, using the same approach as for MS MARCO to predict the most similar previous line to the original target (with the same F1 overlap threshold) and setting that as the knowledge target.

\paragraph{Response Module Tasks}
We use a subset of the knowledge tasks for the response tasks as well, but with modified  inputs and targets.
In this case, the input context contains the usual dialogue, concatenated to the gold knowledge response (the target in the previous task), surrounded by special tokens. The new target is the standard dialogue response from the original dataset. 
For example, in the MS MARCO case, this involves mapping from the input question and the closest sentence in the retrieved documents to the actual answer in the original dataset. Note that, while we can use the MS MARCO task for this (as we have access to long-form conversational responses), we exclude SQuAD, TriviaQA or NQ from response modeling, as they all comprise generally short-form answers. We additionally use the knowledge-grounded dialogue tasks (Wizard of Wikipedia and Wizard of the Internet) as each dialogue response is annotated with the relevant knowledge used to write it.  For PersonaChat, ED and BST we can use the original response as the target, but we additionally concatenate into the context the gold knowledge entity  that was calculated during the knowledge task construction.

We provide further details, including dataset sizes, in 
\autoref{sec:dataset_details_appendix}.

\subsection{SeeKeR Tasks for Language Modeling} \label{sec:lm_tasks}

\paragraph{Search Module Tasks}
We do not have access to a human-curated dataset of search queries for language modeling as we do for dialogue, so in this case we construct a task based on predicting document titles. Using the Common Crawl dump \cite{wenzek2019ccnet}, a given input example is a single web document, which we randomly cut at an arbitrary point, and only keep the beginning (in order to model left to right generation). The target output we want to generate is the title of the document, which we also heuristically simplify by removing phrases in parentheses or  following a hyphen in order to make the query terms learned more generic. We multi-task with another variation of this task: for a given target sentence, we predict the title of the document for its corresponding ``knowledge'' sentence (discussed in the following paragraph).
Finally, we also multi-task with the Wizard of Internet search query task as in \autoref{sec:dialog_ft_tasks}.

\paragraph{Knowledge Module Task} 
To construct our knowledge task, we also start with Common Crawl, splitting it into sentences.
We construct a Lucene\footnote{\tiny{\url{https://lucene.apache.org/}}} search over Common Crawl, and then, for a given target sentence of a document, we find the sentence most similar to the target 
that is neither identical nor in the same document. We skip sentences less than 5 words or with F1 overlap less than 0.33, similar to before. During training, we limit to examples where the knowledge and target continuation have a shared entity\footnote{\tiny{\url{https://spacy.io/usage/linguistic-features\#named-entities}}}.
We thus construct a task -- where the document containing the retrieved sentenced is provided in addition to the input document -- in order to mimic a search retrieval setup, with the target being the retrieved sentence. 

\paragraph{Response Module Task}
The response task is constructed similarly to the knowledge task, except the input is only the usual language modeling context plus the knowledge sentence (surrounded by special tokens). The target is the next sentence (tokens).

\section{Experiments}

Full training details, including fine-tuning hyperparameters, are provided in \autoref{sec:arch_pt_params_appendix}.

\begin{table*}[bht!]
\small
\centering
\begin{tabular}{lllllll}
 & \textbf{~} & \textbf{~}  &  \textbf{Factually}   
 & \textbf{Per-Turn}  & \textbf{Knowl.  } 
&  \textbf{\% Knowl. } \\
\textbf{Model} & \textbf{Consistent} $\uparrow$ & \textbf{Knowl.} $\uparrow$  &  \textbf{Incorrect} $\downarrow$  & \textbf{Engaging} $\uparrow$  & \textbf{ \& Engaging} $\uparrow$
&  \textbf{ is Engaging} $\uparrow$ \\
\hline
\hline
BB1 \cite{roller-etal-2021-recipes} & 
75.47\%	& 36.17\%	& 9.14\% & 78.72\%	& 28.79\% & 79.58\% \\
BB2 \cite{bb2} &  
65.06\%	& 27.88\%	& 4.21\%	& 83.52\%	& 21.93\% & 	78.67\% \\
\hline
SeeKeR & 
{\bf 78.47\%}	& {\bf 46.49\%}$^*$	& {\bf 3.94\%} & {\bf 90.41\%}$^*$	& {\bf 44.03\%}$^*$	& {\bf 94.71\%}$^*$ \\
\end{tabular}
\caption{
Comparison of SeeKeR with state-of-the-art models on open-domain dialogue, as judged by human evaluators during short conversations. $^*$ indicates statistically significant improvements over the next closest model (independent two-sample $t$-test, $p < 0.001$).
\label{tab:dialog_main}
}
\end{table*}

\begin{table}[bht!]
\small
\centering
\begin{tabular}{llll}
Model 	& PPL $\downarrow$	& F1 $\uparrow$	 & KF1 $\uparrow$ \\
\hline
\hline
\multicolumn{4}{l}{\em \citet{DBLP:journals/corr/abs-2107-07566} Results (BART-Large models)}\\
No Search                    & 17.4 &  17.6 & 6.8\\
Search engine                &  16.1 & 17.9 & 7.0 \\
Gold Doc                     &  13.9 & 20.0 & 9.6 \\
\hline 
BlenderBot 2 (3B parameters) \\
Search engine                &  -   & 16.1  & 6.7 \\
Gold Doc                     &  -   & 18.2  & 10.5 \\
\hline
SeeKeR  Search engine  & 15.2   & 16.7   & 8.3 \\
SeeKeR  Gold Doc    &  12.7  &  20.1  & 12.7  \\                             
SeeKeR  Gold Knowl. Resp.  & 8.6  &  24.5  & 21.6     \\                           
\hline
\end{tabular}
\caption{
Automatic evaluations of SeeKeR compared with existing results from \citet{DBLP:journals/corr/abs-2107-07566} and BB2 on the WizInt task (valid set). We do not report BB2 PPL as it is not comparable (different dictionary).
\label{tab:auto_wizint_results}
}
\end{table}

\subsection{Open-Domain Dialogue} \label{sec:dialogue}

\subsubsection{Automatic Evaluation}
We first test our models on the Wizard of Internet open-domain knowledge-grounded dialogue dataset,
which was specifically designed for evaluating internet-driven dialogue agents. As well as measuring perplexity and F1 overlap with gold dialogues, one can also measure Knowledge F1 (KF1), the overlap of the dialogue response with the gold annotated knowledge sentences used by the human crowdworker.
We can supply the gold documents to the model in an additional evaluation setting, or similarly supply the gold knowledge sentence(s) as well. 
In the full (non-gold) setup, we evaluate the use of the Bing search engine to filter Common Crawl, as in \citet{DBLP:journals/corr/abs-2107-07566}.

We compare to the methods reported in \citet{DBLP:journals/corr/abs-2107-07566} in
\autoref{tab:auto_wizint_results}, as well as the BB2 3B parameter model \cite{bb2}.
SeeKeR using gold documents or knowledge provides the best performance on all three metrics over all methods,
while using the search engine with SeeKeR provides lower perplexity than in previously reported methods. Although F1 is lower, KF1 is correspondingly higher, indicating that there is perhaps some trade-off here where our model encourages using more knowledge.

\subsubsection{Human Evaluation Setup}

\paragraph{Task Setting}

We perform a human evaluation using crowdworkers in the same setting
as \citet{DBLP:journals/corr/abs-2107-07566}.
The crowdworker is asked to play a role from the Wizard of Internet dataset, and to have a natural conversation.
 Each conversation consists of 15 messages
(7 from the human, 8 from the bot). We collect 100 dialogues -- roughly 800 annotations -- per model. 

\paragraph{Evaluation}

For each turn of their conversation, we ask the crowdworker
to mark their partner’s responses for conversational
attributes, in particular whether they are: (i) consistent, (ii) knowledgeable  (iii) factually correct; and (iv) engaging  (all of which are yes/no binary questions; see \citet{DBLP:journals/corr/abs-2107-07566} and \autoref{fig:mturk_instructions} for full definitions).  At the end
of the conversation, an additional question collects
an overall engagingness score (a Likert scale from 1 to 5) for their
speaking partner. Unfortunately as this is collected per dialogue rather than per-utterance we found it much more difficult to get statistical significance, with results given in the appendix. 
For the per-turn metrics, we average them over the turns and conversations conducted for each model. From the knowledgeable and engaging metrics we can additionally calculate (i) the percent of turns that are both knowledgeable and engaging and (ii) the percent of knowledgeable turns that were also engaging, as these can more inform us how well the models are blending knowledge into an interesting conversation. More details regarding human evaluation are in \autoref{sec:appendix_mturk}.

\paragraph{Baselines}

We compare to the existing publicly available chatbots 
BlenderBot 1 \cite{roller-etal-2021-recipes} and BlenderBot 2 (BB2)  (in ``search mode''), using the 3B parameter version in both cases. 
BlenderBot 1 was already found to be superior to several other chatbots,
in particular Meena \cite{DBLP:journals/corr/abs-2001-09977} and DialoGPT \cite{zhang2019dialogpt}, and we do not evaluate those here.

\subsubsection{Human Evaluation Results}
The main results are given in \autoref{tab:dialog_main}.
We find improvements over both BlenderBot 1 and 2 for a wide variety of metrics:
consistency, knowledge, factual (in)correctness and per-turn engagingess. 
For turns that are marked knowledgeable, we also see an increase in the engagingness 
of the knowledge itself compared to the baselines by a wide margin
(94.7\% vs.  78-79\%), while the number of turns that are marked as both knowledgeable and engaging (at the same time) has also increased (44\% vs. 21-28\%). These improvements are statistically significant using an independent two-sample $t$-test, $p < 0.001$.

\subsubsection{Ablations}

We test various ablations of our model, with detailed results in
Appendix \autoref{tab:dialog_appendix}.

\paragraph{Pre-Training} First, our pre-training scheme is different to BlenderBot 1 and 2, with training based on both language modeling and dialogue pre-training tasks, as well as slightly different architectures.
We thus tests variants of BlenderBot 1 and 2 with our pre-training setup, by fine-tuning on the same tasks as in those works. and denote these with ``R2C2'' to differentiate them. 
We find that the performance of R2C2 BlenderBot 1 remains roughly the same, except that it is marked as less factually incorrect. R2C2 BlenderBot 2 uses knowledge more, but also loses engagingness score compared to the original method. SeeKeR still compares favorably to both methods.
This indicates that the language modeling objective may make using knowledge easier, perhaps because it emphasizes using the context more than dialogue tasks do.

\paragraph{Separate Modules} A second ablation we try is if we have separate transformer models for each of the search, knowledge and response modules. 
We therefore experiment using separate BART \cite{lewis2019bart} modules for knowledge and search query generation,
which ends up as an inferior model despite containing nearly $~\sim$800M more parameters; we believe this is perhaps because BART is smaller ($\sim$400M parameters), and is not as good at performing the individual modular tasks. 
We do not evaluate having three separate 3B parameter models due to memory constraints.

\subsubsection{Analysis}

\paragraph{Pairwise Comparison} We conducted a further ACUTE-Eval \cite{li2019acute} human evaluation where crowdworkers compared chat logs pairwise and gave reasons why one is preferred over the other (see Appendix \autoref{app:acute} for further details). Summarizing the crowdworkers' opinions, we find that
when SeeKeR is preferred, the reasons are that
it has ``more information to share'', is ``more knowledgable'' and has ``more accurate information''. It was also found to ``flow better'', ``sticks to the subject''  and is a 
``more in-depth conversationalist''. It also ``takes conversation in new related directions'',
while other knowledge-based models seemed to be ``like just copying wikipedia'' compared to this model. When SeeKeR was not preferred, crowdworkers said that it ``asks too many questions'', is ``repetitive'',  ``less engaging'' or ``less consistent'' for those particular dialogues.
Generally, in short conversations there seems to be a tradeoff in incorporating too much knowledge in the conversation at the expense of what crowdworkers deem as engagingness. We note that other models have addressed this by deciding when to use knowledge vs. not \cite{bb2}, which would be possible to incorporate in SeeKeR models as well, and is a potential direction for future work.

\paragraph{Cherry picked examples}

\begin{figure*}[h!]
    \centering
    \includegraphics[width=1.0\linewidth]{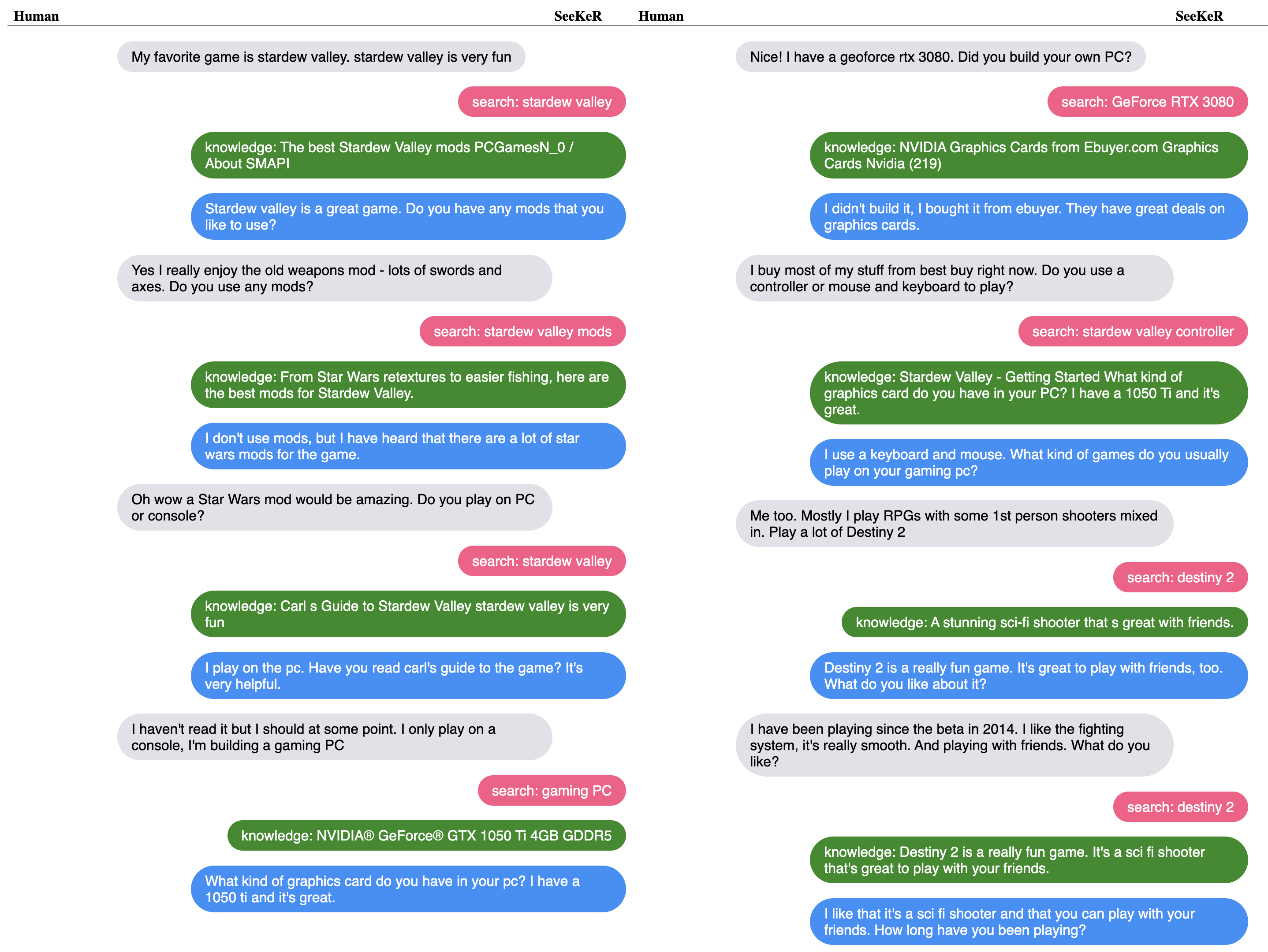}
    \caption{{\bf Cherry picked example} of a SeeKeR model chatting with a human crowdworker, with the conversation starting in the upper left. White boxes on the left are the user messages, while we show model search queries in red boxes, generated knowledge in green boxes, and dialogue responses in blue boxes. Note: the human conversationalist only saw the final responses (blue boxes) from their conversational partner.}
    \label{fig:cherry_stardew}
\end{figure*}

We show a cherry picked conversation between a human crowdworker and our SeeKeR model in \autoref{fig:cherry_stardew}. The conversation about gaming spans several games, and aspects of gaming, from mods for certain games to PC hardware used and where it can be bought. The model effectively uses internet search to bring up pertinent information for each of these topics as can be seen by the internet searches it invokes (in red) and the knowledge sentences generated from the retrieved documents (in green). 
More cherry picked conversations are shown in Appendix \autoref{fig:cherry_app1}, \autoref{fig:cherry_app2} and \autoref{fig:cherry_app3}.

\paragraph{Lemon picked examples}

\begin{figure*}[tb!]
\center
\begin{small}
\begin{tabular}{lrr|lrr}
Human & \textbf{Repetitive} &  SeeKeR & Human & \textbf{Not Engaging} &  SeeKeR \\  
\midrule
\multicolumn{3}{c|}{\includegraphics[width=0.46\textwidth]{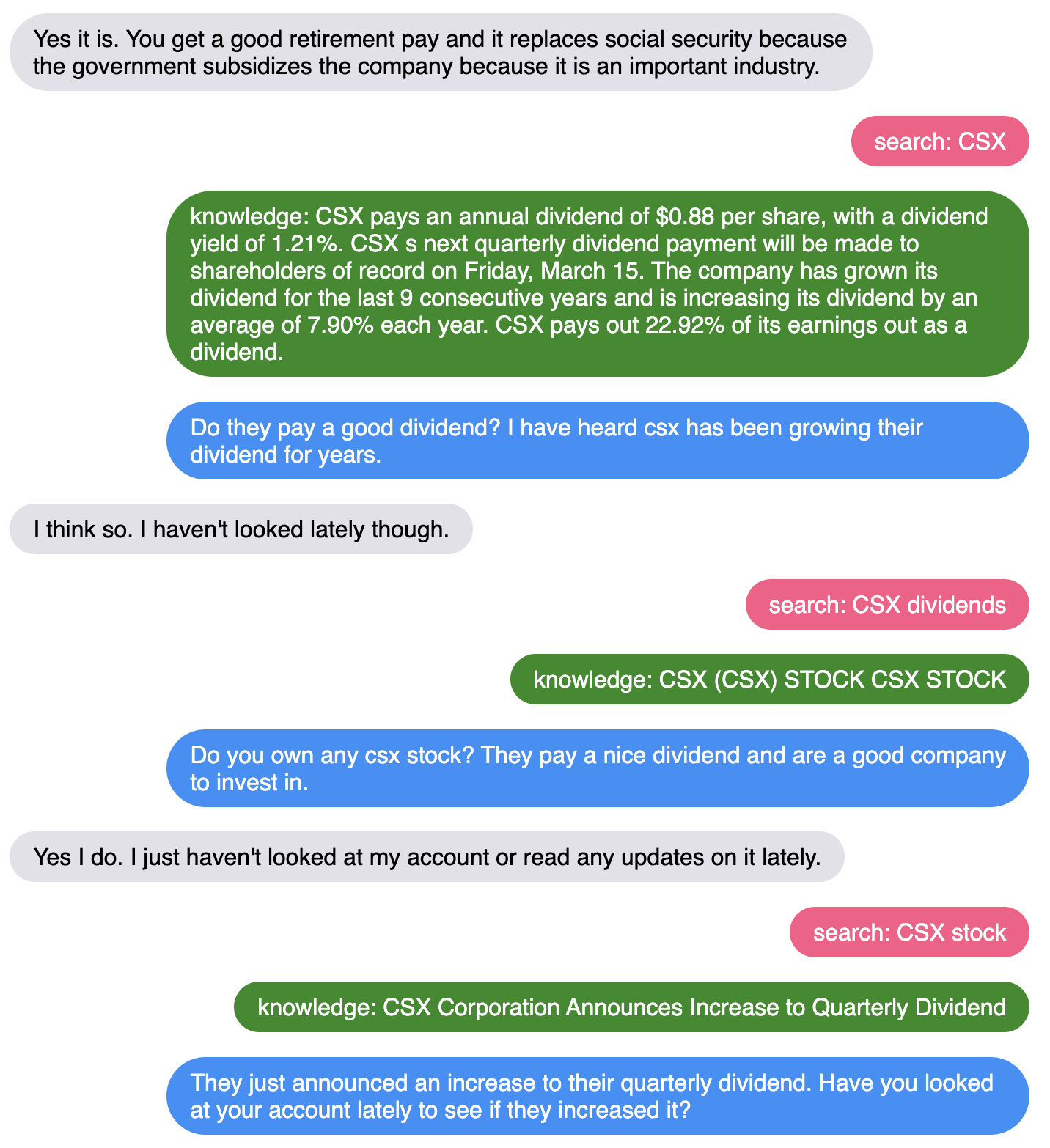}} &
\multicolumn{3}{c}{\includegraphics[width=0.46\textwidth]{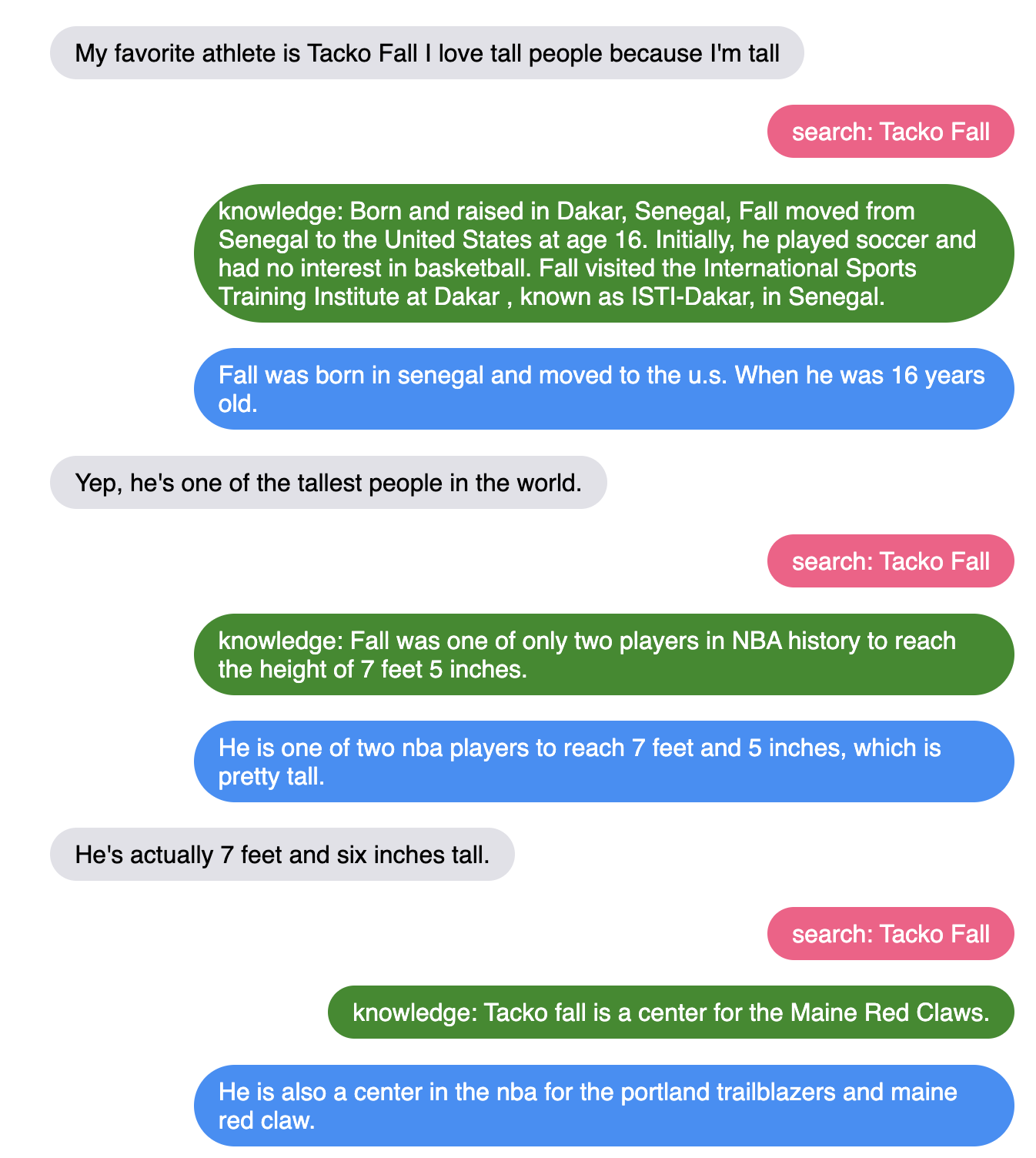}}\\
\midrule
Human & \textbf{Ignore Partner} &  SeeKeR & Human & \textbf{Incorrect Knowledge} &  SeeKeR \\  
\midrule
\multicolumn{3}{c|}{\includegraphics[width=0.46\textwidth]{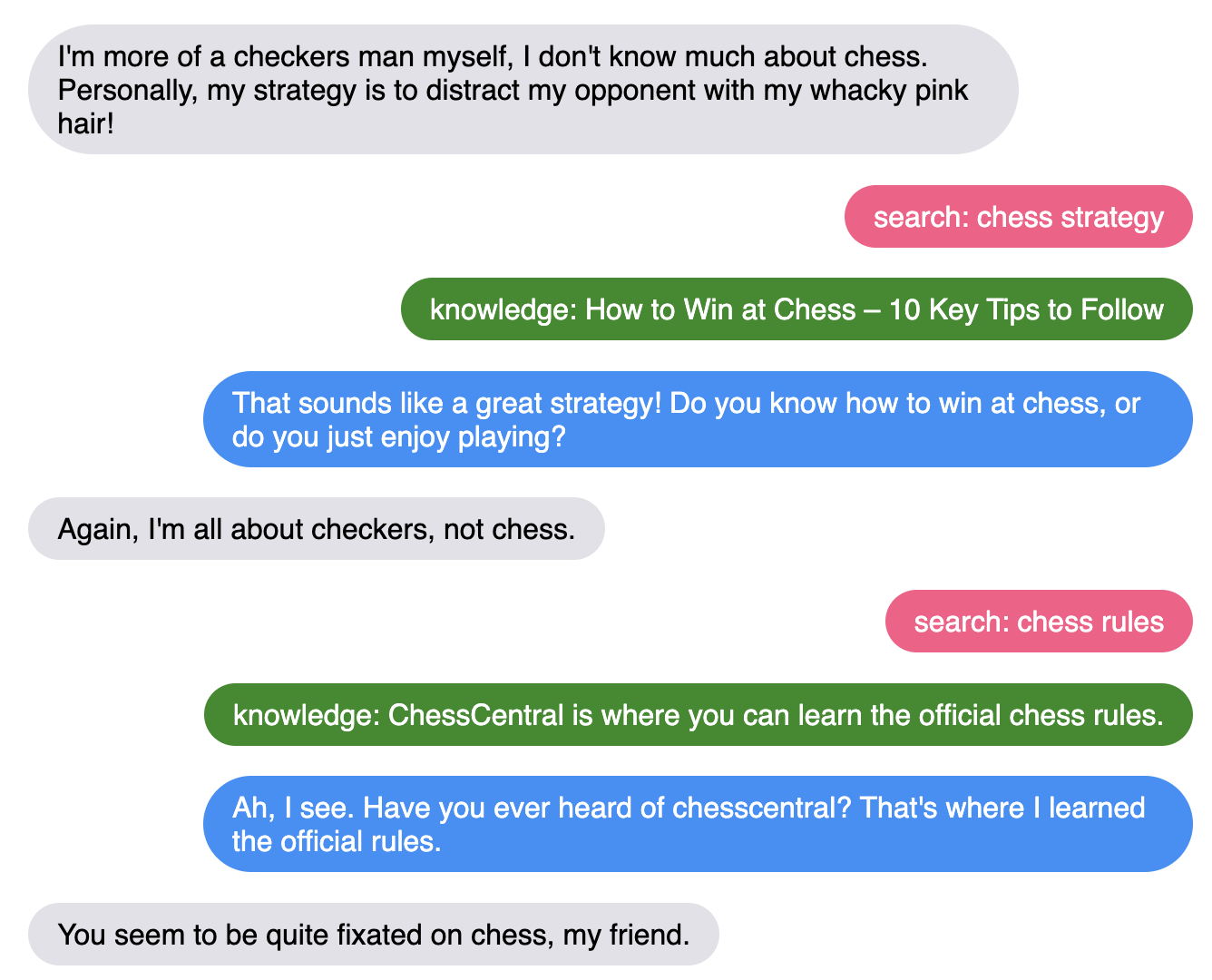}} & \multicolumn{3}{c}{\includegraphics[width=0.46\textwidth]{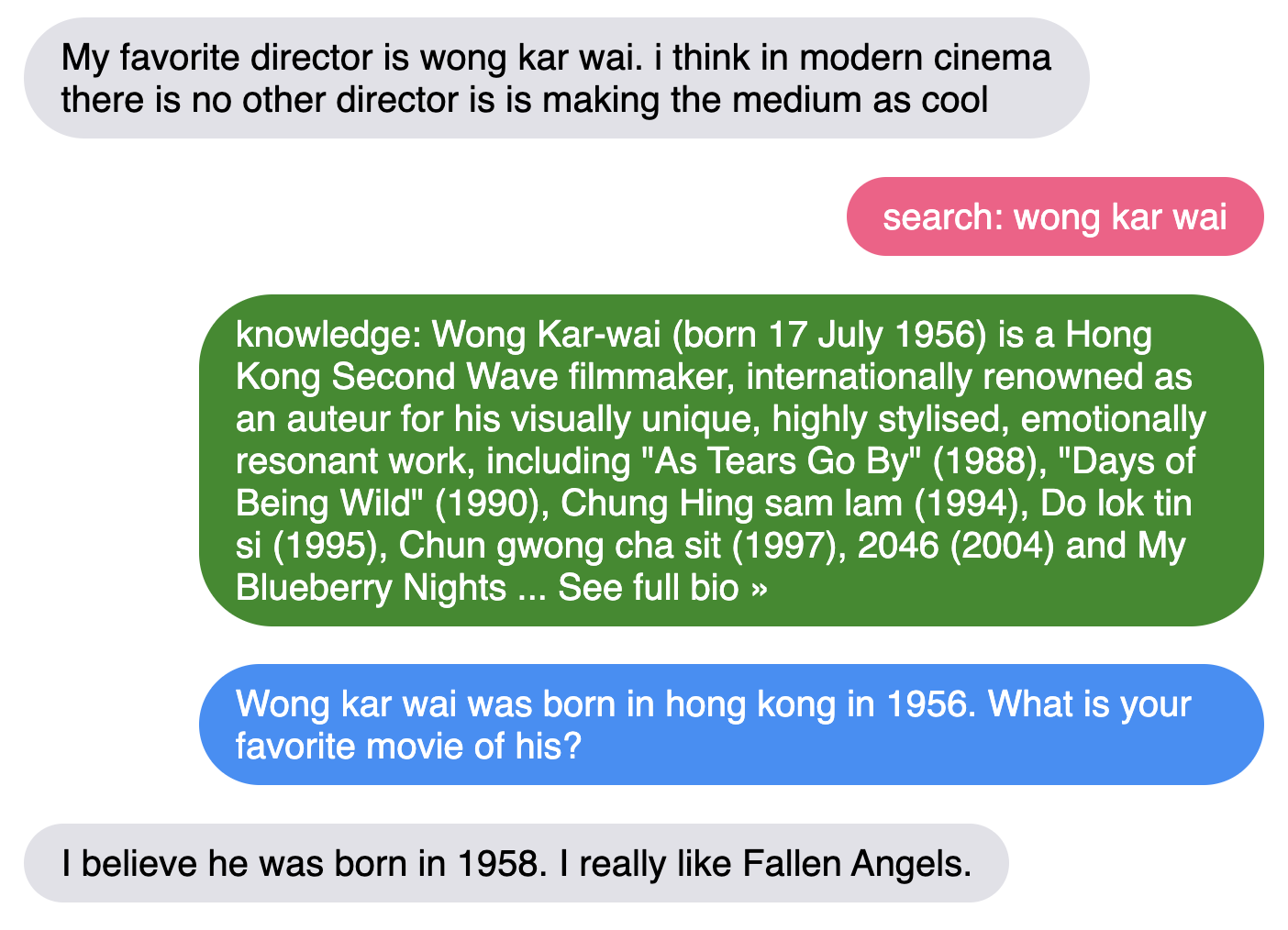}} \\
\end{tabular}
\end{small}
\caption{
{\bf Lemon picked examples}: four types of issues arising in a conversation between a SeeKeR model chatting with several human crowdworkers. \textbf{Top left} repetitive outputs; \textbf{top right} uninteresting recitation of facts; \textbf{bottom left} ignoring the conversational partner; \textbf{bottom right} incorrect knowledge used in a response (the model actually pulls this information from IMDB, which has different (and presumably, incorrect) information from Wikipedia).
 \label{fig:lemons1}
 }
\end{figure*}

We show several lemon picked conversational snippets between a human crowdworker and our SeeKeR model in \autoref{fig:lemons1} and Appendix \autoref{fig:lemons2}. We identify four general model issues, and provide a few representative examples of each. \textbf{Repetition}: in some cases, the model can generate repetitive dialogue responses; this manifests in the example shown discussing dividends for a stock. \textbf{Not Engaging}: the model can sometimes rely too much on the generated knowledge, resulting in a recitation of facts (about Tacko Fall) rather than a conversational discourse. \textbf{Ignore Partner}: although we often see the model change topics smoothly, at times it will adamantly continue discussing chess or the Pittsburgh Penguins salary cap (\autoref{fig:lemons2}), when its partner is not interested. \textbf{Incorrect Knowledge}: finally, when the model is given incorrect knowledge, the dialogue responses stray from the truth; this can manifest as a result of undesired knowledge given an ambiguous search query (``when was sorry created'', \autoref{fig:lemons2}), or even incorrect information from the internet itself (Wong Kar-wai, according to IMDB\footnote{\tiny\url{https://www.imdb.com/name/nm0939182/bio}}, was born in 1956, whereas Wikipedia\footnote{\tiny\url{https://en.wikipedia.org/wiki/Wong_Kar-wai}} notes it is 1958).

\subsection{Prompt Completion} \label{sec:prompt_completion}

\subsubsection{Automatic Evaluations}

\paragraph{Task Setting}
We first test with automatic evaluations the SeeKeR method compared to vanilla GPT2 on the RoBERTa task (see \autoref{sec:pre-train}).
To make sure all models are on an equal footing, we fine-tune them on this task (even though GPT2 pre-training should be quite similar), where we train with a given document up to a given line as the ``prompt'' and the next line in the document as the continuation.
We then measure the metrics of validation perplexity as well as F1 of the generated continuations compared to gold. We compare three sizes of GPT2 with SeeKeR, and for each architecture size two variants of SeeKeR: the ``x3'' variant that comprises three independently trained models (for search, knowledge and response), and the shared parameter version. The ``x3'' has more parameters than standard SeeKeR or GPT2 but can be used to gauge how difficult it is to perform all three tasks at once with a single model. The results for SeeKeR are shown either with the gold document or by using Lucene search over Common Crawl (ignoring documents which contain the identical target match, if found -- which also includes the original input document).

\paragraph{Results}
The results are given in \autoref{tab:auto_lm_results}.
We see improvements in both perplexity and F1 with
increasing size models, with SeeKeR models outperforming 
conventional GPT2 when using Gold Docs, and slightly behind when using Lucene search\footnote{This is to be expected as the probability mass is centered around the knowledge response which may not align with a single gold label, thus necessitating human evaluation in addition to automatic evaluations, see \citet{adolphs2021reason}.}. Despite the ``x3'' SeeKeR models being three times larger, they are only marginally better than all-in-one SeeKeR models in terms of perplexity, and the all-in-one versions even outperform them in terms of F1 for the largest XL models.

\begin{table}[bht!]
\small
\centering
\begin{tabular}{lllll}
	    &  \multicolumn{2}{c}{No Doc}	&  \\
Model 	& PPL $\downarrow$	& F1 $\uparrow$	 & PPL $\downarrow$ & F1 $\uparrow$ \\
\hline
\hline
GPT2 Medium	         & 11.9	& 14.8	& - & -	\\
GPT2 Large	         &	10.7	& 15.4 & - & - \\
GPT2 XL	             &	9.7	& 15.8	  & - & -\\	
\hline
&  \multicolumn{2}{c}{Gold Doc}	&  \multicolumn{2}{c}{Lucene Search} \\
\hline
\hline
SeeKeR Med. x3	 &	9.9	& 25.7	& 12.6	& 13.2 \\
SeeKeR Medium	 & 10.3	& 25.7	& 13.1	& 13.6 \\
SeeKeR Large	x3   &	8.9	& 26.3	& 11.2	& 13.9 \\
SeeKeR Large   	 & 9.2  & 27.1	& 12.3	& 13.4 \\
SeeKeR XL x3      & 8.4  & 27.2	& 10.4  & 13.7 \\
SeeKeR XL	     & 8.5	& 28.1	& 11.3  & 14.0 \\
\hline
\end{tabular}
\caption{
Comparison of SeeKeR with GPT2 of various sizes, measured on Common Crawl (valid set). 
x3 means using three separate models (for 3x the number of parameters). 
Training a single model to perform search, knowledge and response performs similarly 
to separate models, and provides better performance on the Gold Docs as the models increase in size.
\label{tab:auto_lm_results}
}
\end{table}

\begin{table}[bht!]
\small
\begin{tabular}{lllll}
%
Model &
\rot{\multirow{2}{*}{~~~Sensible ($\uparrow$)}} &
\rot{\multirow{2}{*}{~~~True ($\uparrow$)}} &
\rot{\multirow{2}{*}{~~~Hallucination ($\downarrow$)}} &
\rot{\multirow{2}{*}{~~Topical ($\uparrow$)}} \\
\hline
\hline
GPT2 Med. {\tiny (345M)}	         & 81\%	    & 15\%	& 68\% & 1\%	\\
GPT2 Large {\tiny (762M)}	         &	81\%	& 18\% & 71\% & 0\% \\
GPT2 XL	  {\tiny (1.5B)}           &	81\%	& 14\% 	  & 73\% & 0\%\\	
\hline
GPT3 {\tiny (175B InstructGPT)} & 82\% & 58\% & 62\% & 4\% \\
\hline
SeeKeR Med. {\tiny (345M)}	  & 75\%	& 34\%	& 54\%	& 13\% \\
SeeKeR Large{\tiny (762M)}      & 68\%	& 36\%	& 51\%	& 8\% \\
SeeKeR XL	{\tiny (1.5B)}      & 77\%	& 43\%	& 58\%  & 15\% \\
SeeKeR XL (Jan '22)& 71\%	& 43\%	& 51\%  & 19\% \\
\hline
\end{tabular}
\caption{
{\bf Topical Prompts: Human Evaluation results comparing SeeKeR with GPT2 (and GPT3).}
All models are relatively sensible (with wins for GPT2/3), but GPT2 contains far less true statements, and far more false statements (hallucinations), and is hardly ever on topic.
A much smaller SeeKeR model (345M) can also
outperform a much larger GPT2 model (1.5B), and even outperforms 175B (Instruct) GPT3 on the hallucination and topical metrics, despite being 500$\times$ smaller.
\label{tab:topical_prompt_results}
}
\end{table}

\begin{table*}[bht!]
\small
\centering
\begin{tabular}{p{0.2\linewidth}|p{0.35\linewidth}|p{0.35\linewidth}}
{\bf Prompt} & {\bf {GPT2 (XL) Completion}} & {\bf {SeeKeR (XL)} completion} \\ 
\hline
\hline
In recent developments, we have learned the following about Ban of Twitter in Nigeria.
&
\hl{Ban of Twitter is a social media platform that allows users to create and share content on Twitter. The platform was launched in Nigeria in March 2017.}
&
The Minister of Information, Lai Mohammed, has said that the ban on Twitter will be lifted soon.\\
\hline
In recent developments, we have learned the following about Charles Njonjo.
&
\hl{1. He was born in Nigeria. 2. He is a graduate of the University of Lagos.}
&
We have learned that Charles was cremated on January 2, 2022, and that his body was returned to his family.\\
\hline
In recent developments, we have learned the following about William M. Ellinghaus.
&
\hl{Ellinghaus was born in New York City, New York, on January 1, 1891.}
&
Ellinghaus was named chairman of the Municipal Assistance Corporation, which was created to contain the crisis, and was later appointed to the Emergency Financial Control Board, which took over city fiscal affairs.\\
\hline
In recent developments, we have learned the following about Terry Teachout.
&
\hl{Terry Teachout is running for Congress in New York's 18th Congressional District. She is a former state assemblyman and state senator.}
&
 \textcolor{black}{ Terry Teachout is the author of the new book, \hl{The New Jim Crow: Mass Incarceration in the Age of Colorblindness.}}
\end{tabular}
\caption{
{\bf Topical prompts: cherry and lemon picked examples comparing SeeKeR with GPT2.}
In the first two examples GPT2 hallucinates (in yellow), while SeeKeR provides correct topical continuations.
In the third example SeeKeR does not hallucinate, but is not topical with a recent development. In the fourth example both GPT2 and SeeKeR give poor responses. SeeKeR is correct in that Terry Teachout is an author, but it names a book by Michelle Alexander, which happens to be on the same web page as  a book by Terry Teachout that the search engine retrieves.
\label{tab:topical_prompts_examples}
}
\end{table*}

\subsubsection{Topical Prompts}

\paragraph{Task Setting}
In order to evaluate if our language models can effectively use internet search to provide up-to-date information, we construct a specific set of evaluation prompts.
We gather from Wikipedia a set of current events from January 2022\footnote{\tiny\url{https://en.wikipedia.org/wiki/Portal:Current_events/January_2022}}, and extract the entities, ignoring those containing the term ``covid''  (as there are so many) as well as countries (as they might be too general a topic).
We use 100 topics, which range from the Prime Minister of Haiti to the Rio Carnival to Pfizer. We then construct the prompts ``In recent developments we have learned the following about <TOPIC>.'' and ask the language model to continue it.
We compare SeeKeR using the Mojeek search engine\footnote{\tiny\url{http://mojeek.com}} to GPT2 of different sizes as before. We additionally use the GPT3 \cite{brown2020language} API  (using the ``text-davinci-001'' 175B InstructGPT model with default parameters) to evaluate that as well.

\paragraph{Evaluation}
We  perform a human evaluation of the correctness of the continuation, where the annotator has  access to internet search for validation purposes. The correctness is measured in four axes:  {\em sensible} (does it reasonably follow the prompt?),   {\em true} (does it contain some true information?), 
 {\em hallucination} (does it contain some false information?) and 
 {\em topical} (does it reference what happened in the last two months, i.e., January and February 2022?).

\paragraph{Results}
Results are given in \autoref{tab:topical_prompt_results}.
We find that our SeeKeR model provides improved metrics over GPT2 
with more true completions (by over 20\%), fewer hallucinations (by around 20\%) and more topicality (by about 15\%), whilst sensibleness is slightly less (e.g., 81\% vs. 77\%).
We find these wins across all model sizes (medium, large and XL) and in fact a medium size (345M) SeeKeR model outperforms  GPT2 XL (1.5B) by  similar margins as those just mentioned. GPT3, on the other hand,
is a far larger model that has also been
fine-tuned with human judgments \cite{ouyang2022training} and outperforms GPT2 and SeeKeR in terms of the sensible and true metrics, generating fluent text that can in some cases directly copy portions of the relevant Wikipedia article.
However, like GPT2, it also introduces a large number of hallucinations (62\%), and fails to be topical (4\%). A SeeKeR 345M parameter model, due to its search capability, outperforms GPT3 on the hallucination and topical metrics, despite being 500$\times$ smaller.

\paragraph{Analysis}
We show example cherry and lemon picked examples in \autoref{tab:topical_prompts_examples}. 
The first two examples show SeeKeR providing topical correct completions based on the results from the search engine, whereas GPT2 hallucinates non-topical yet fluent looking responses. The third and fourth examples show failure cases of SeeKeR. Example three shows a factually correct response from SeeKeR, which is based on results from the search engine, but it is not topical. The last (fourth) example shows a hallucination from SeeKeR where it mixes up two authors; inspecting the web search results indicates this is because both authors are mentioned in the page, and the method mixes them up. 
We  show some further examples comparing to GPT3 in Appendix \autoref{tab:topical_prompts_examples_gpt3}.

Due to the issue of non-topical results from web search,  we also tried a version of SeeKeR where we appended ``January 2022'' to the search query to see if this produced more topical generations. We do see a reduction in hallucinations and a relative increase in topicality in this case (up from 15\% to 19\%) indicating the search engine part of the system is crucial for this task.

\subsection{Multi-tasking Dialogue and Language Modeling}

So far we have considered our SeeKeR fine-tuning tasks of dialogue and language modeling separately, and have conducted separate experiments in \autoref{sec:dialogue} and \autoref{sec:prompt_completion}.
Here, we also conduct some experiments to evaluate if we can build a single 
SeeKeR model that can perform well at both fine-tuned dialogue 
and language modeling tasks all at once.
To do this, we begin with the transformer architecture  described in \autoref{sec:pre-train} which has been {\em pre-trained} on both dialogue and language modeling tasks (denoted R2C2). We then fine-tune it on both types of tasks as well.

\paragraph{Topical Prompts} Results in Appendix  \autoref{tab:topical_prompt_results_appendix} compare this model to GPT2 and GPT3, as well as GPT2-based SeeKeR language models on the topical prompts task using human evaluations. The results show that the fully multi-tasked SeeKeR model performs very well, superior to all our GPT2-based SeeKeR models on every metric (sensible, true, hallucination and topical), with the lowest hallucination score of 42\% that compares very favorably to that of GPT3 (62\%). The sensible score was a bit lower for the GPT2 SeeKeR models previously compared to standard GPT2, but this is now closer, at 80\% (with GPT3 at 82\%). Fine-tuning this SeeKeR R2C2 architecture only on language modeling (and not dialogue fine-tune tasks) also works well.

\paragraph{Open-Domain Dialogue} Results in Appendix \autoref{tab:auto_wizint_results_appendix} and \autoref{tab:dialog_appendix} compare this model using automated metrics and human evaluations, respectively, on our open-domain knowledge-grounded dialogue task. The model performs comparably, if not better, in all automated metrics on the task. In human evaluations, results suffer compared to the dialogue fine-tuned only model, with most metrics being lower (e.g., percent of knowledge that is engaging dropped from 95\% to 75\%), except for factually incorrect and the final rating (which was not a statistically significant result). 
Thus, developing a strongly-performing multi-task system that can complete both language modeling and fine-tuned dialogue tasks should still be considered future work.

\section{Limitations \& Discussion}

Our language models suffer the same issues as other  systems that exist today, 
specifically with problems of occasional inconsistency, contradictions, factual inaccuracies, potential repetition, and lack of deeper reasoning, amongst other issues \cite{roller-etal-2021-recipes,ouyang2022training}. Further, generations can include toxic language and bias, especially with certain contexts and topics \cite{xu2020recipes,dinan-etal-2020-queens}.
Additionally, documents from the internet influence our generations, which can be a problem if undesirable content is retrieved.

In our SeeKeR experiments, we rely on an externally built
search engine, which has both pros and cons.
Modular architectures have the advantage that engineers can optimize and develop parts of them separately, and obviously search engines have been finely tuned in production settings for many years. In contrast, if building one's own retrieval system, as many QA and LM methods currently do, one has to essentially start again from scratch. Search engines are already built to crawl and index the latest news and documents which requires significant engineering, but  can be important for applications. Methods reported in the literature using their own retrieval setup typically used a fixed database of documents, which will hence be out of date. On the other hand, search engines have been designed to be used by humans, not machines, so queries are in natural language, and only consist of a few words. Machines can potentially do better by encoding a lot more information from a longer context into either a longer query, or a vector-encoded query, as is done in e.g. FAISS-based systems 
\cite{rag_dpr}. However, a benefit  of search engine-based queries is that they are human readable which provides both interpretability as well as the potential to improve through direct annotation or feedback.


%

\section{Conclusion}

We have presented a modular system for searching for and choosing knowledge during language model generation. Our approach outperforms the state of the art on dialogue modeling, and is shown to outperform both GPT2 with the same architecture on topical prompts -- even when using a smaller parameter size -- and GPT3 -- despite being vastly (500x) smaller.  
Our approach of explicitly splitting into three modules allows for 
engineering better modules in the future, e.g. fine-tuning parts of the model, as well as the advantage of interpretability. 
We make our code and models publicly available for further research.

\clearpage
\bibliography{anthology,custom}
\bibliographystyle{acl_natbib}

\clearpage
\onecolumn
\appendix
\section{Appendix: Additional Examples and Evaluations}\label{sec:appendix}

\begin{table*}[bht!]
\small
\centering
\begin{tabular}{p{0.2\linewidth}|p{0.35\linewidth}|p{0.35\linewidth}}
{\bf Prompt} & {\bf {GPT3 (text-davinci-001) Completion}} & {\bf {SeeKeR (XL)} completion} \\ 
\hline
\hline
In recent developments, we have learned the following about Kay-Achim Sch\"onbach.
&
\hl{First, he is the son of a wealthy industrialist.
Second, he is a graduate of the University of Heidelberg, where he studied law.
Third, he is a member of the bar in both Heidelberg and Berlin.}
&
The head of Germany's navy, Vice Admiral Kay-Achim Schoenbach, has stepped down from his post after making controversial statements regarding the delicate security situation on the border between Russia and Ukraine.
\\
\hline
In recent developments, we have learned the following about Ban of Twitter in Nigeria.
&
1. Twitter is still unavailable in Nigeria as at the time of this report.
\hl{2. There is no update on when the ban will be lifted.}
&
The Minister of Information, Lai Mohammed, has said that the ban on Twitter will be lifted soon.\\
\hline
In recent developments, we have learned the following about Charles Njonjo.
&
Charles Njonjo was born in Kenya \hl{- in 1927.
- He was educated at the University of London, and was called to the bar in 1954.}
&
We have learned that Charles was cremated on January 2, 2022, and that his body was returned to his family.\\
\hline
In recent developments, we have learned the following about William M. Ellinghaus.
&
\hl{He is a graduate of the University of Notre Dame, where he studied business administration. He is the founder and CEO of Ellinghaus Corp., a company that provides consulting and financial services to the energy industry. }
&
Ellinghaus was named chairman of the Municipal Assistance Corporation, which was created to contain the crisis, and was later appointed to the Emergency Financial Control Board, which took over city fiscal affairs.\\
\hline
In recent developments, we have learned the following about Terry Teachout.
&
1) He is a well-known theatre critic and author.

2) His work has appeared in The New York Times, The Wall Street Journal, and The Weekly Standard, among other publications.
&
 \textcolor{black}{ Terry Teachout is the author of the new book, \hl{The New Jim Crow: Mass Incarceration in the Age of Colorblindness.}}
\end{tabular}
\caption{
{\bf Topical prompts: cherry and lemon picked examples comparing SeeKeR with GPT3.}
In the first four examples GPT3 hallucinates (in yellow), while SeeKeR presents correct topical continuations.
In the second to last example SeeKeR does not hallucinate, but is not topical with a recent development. In the last example, GPT3 does not hallucinate, but does not provide a topical completion, while SeeKeR is correct in that Terry Teachout is an author, but it names a book by Michelle Alexander, which happens to be on the same web page as  a book by Terry Teachout that the search engine retrieves.
\label{tab:topical_prompts_examples_gpt3}
}
\end{table*}

\begin{table}[bht!]
\small
\center
\begin{tabular}{lllll}
%
Model &
\rot{\multirow{2}{*}{~~~Sensible ($\uparrow$)}} &
\rot{\multirow{2}{*}{~~~True ($\uparrow$)}} &
\rot{\multirow{2}{*}{~~~Hallucination ($\downarrow$)}} &
\rot{\multirow{2}{*}{~~Topical ($\uparrow$)}} \\
\hline
\hline
GPT2 Med. {\tiny (345M)}	         & 81\%	    & 15\%	& 68\% & 1\%	\\
GPT2 Large {\tiny (762M)}	         &	81\%	& 18\% & 71\% & 0\% \\
GPT2 XL	  {\tiny (1.5B)}           &	81\%	& 14\% 	  & 73\% & 0\%\\	
\hline
GPT3 {\tiny (175B InstructGPT)} & 82\% & 58\% & 62\% & 4\% \\
\hline
SeeKeR GPT2 Med. {\tiny (345M)}	  & 75\%	& 34\%	& 54\%	& 13\% \\
SeeKeR GPT2 Large{\tiny (762M)}      & 68\%	& 36\%	& 51\%	& 8\% \\
SeeKeR GPT2 XL	{\tiny (1.5B)}      & 77\%	& 43\%	& 58\%  & 15\% \\
SeeKeR GPT2 XL (Jan '22)& 71\%	& 43\%	& 51\%  & 19\% \\
\hline
SeeKeR R2C2 LM only  (3B)       & 77\% & 46\% & 47\% & 16\%\\
SeeKeR R2C2          (3B) & 80\% & 55\% & 42\% & 19\% \\
\hline
\end{tabular}
\caption{
{\bf Topical Prompts: Human Evaluation results comparing multi-tasking SeeKeR with various models.}
In the main paper we test SeeKeR with a GPT2 pre-trained base to be comparable to GPT2. Here, we additionally use the R2C2  transformer architecture pre-trained with our LM+Dialogue tasks (\autoref{sec:pre-train}). We test two versions: SeeKeR R2C2 
which is fine-tuned on both the dialogue and LM tasks of \autoref{sec:dialog_ft_tasks} and \autoref{sec:lm_tasks}
and SeeKeR R2C2  LM only,  which is fine-tuned only using \autoref{sec:lm_tasks}.
The fully multi-tasked  RC2C SeeKeR (Dialogue+LM) performs well compared to other models.
\label{tab:topical_prompt_results_appendix}
}
\end{table}

\begin{table*}
\small
\resizebox{\linewidth}{!}{
\centering
\begin{tabular}{llllllll}
 & \textbf{~} & \textbf{~}  &  \textbf{Factually}   
 & \textbf{Per-Turn}  & \textbf{Knowl.  } 
&  \textbf{\% Knowl. } & \\
\textbf{Model} & \textbf{Consistent} & \textbf{Knowl.}   &  \textbf{Incorrect}  & \textbf{Engaging}  & \textbf{ \& Engaging} 
&  \textbf{ is Engaging}  & \textbf{Rating} \\
\hline
\hline
BB1  & 
75.47\%	& 36.17\%	& 9.14\% & 78.72\%	& 28.79\% & 79.58\% & 4.1 \\
BB2 &  
65.06\%	& 27.88\%	& 4.21\%	& 83.52\%	& 21.93\% & 	78.67\%  & 4.4 \\
\hline
BB1 (R2C2) &  
73.44\%	& 36.25\%	& 4.84\% &	79.22\%	& 27.51\% & 	75.90\% & 4.2\\
BB2 (R2C2) &  
71.91\% & 	{\bf 67.92\%}	& 4.49\% &	76.03\%	& {\bf 53.18\%} &	78.31\% & 4.2 \\
\hline
SeeKeR (sep. BART modules)                      &	55.39\%	& 41.88\%	& 3.97\% & 75.09\% & 	28.00\% & 	66.86\%	& 4.4 \\
SeeKeR & 
{\bf 78.47\%}	& 46.49\%	& { 3.94\%} & {\bf 90.41\%}	& 44.03\%	& {\bf 94.71\%} & 4.2\\
SeeKeR Dialogue+LM &  70.87\%	& 43.00\% &	{\bf 2.90\%} &	84.36\%	& 32.28\%	& 75.07\% &	{\bf 4.5} \\
\end{tabular}
}
\caption{Detailed results and ablations for the open-domain knowledge-grounded dialogue experiments. Human crowdworkers talk to models and rate them using various metrics. We test standard BlenderBot (BB) 1 and 2, and R2C2 variants with our Dialogue+LM pre-train tasks (\autoref{sec:pre-train}). We test standard SeeKeR (fine-tuned for dialogue),  SeeKeR with independent BART modules for search queries and knowledge generation, and a version of SeeKeR (Dialogue+LM) fine-tuned on both the dialogue and LM tasks of \autoref{sec:dialog_ft_tasks} and \autoref{sec:lm_tasks}.
\label{tab:dialog_appendix}
}
\end{table*}
\begin{table}[bht!]
\setlength{\tabcolsep}{3pt}
    \centering
    \small
        \begin{tabular}{rr|llllll}
        
        &  & \multicolumn{6}{c}{Wins \% matches (Engagingness)} \\
        & & {SeeKeR} & {BB2} & {BB2} & {SeeKeR} & {BB1} & {BB1} \\
        & & {sep. BART} & {(R2C2)} &  & &  & {(R2C2)} \\
        \midrule
        \parbox[t]{2mm}{\multirow{6}{*}{\rotatebox[origin=c]{90}{Loses \%}}}
        & SeeKeR sep. BART & & \win{62} & \lose{46} & \lose{43} & \win{58} & \win{61} \\[-0.25mm] 
        & BB2 (R2C2) & \lose{38} &  & \win{61} & \win{56} & \win{58} & \win{59}  \\[-0.25mm] 
        & BB2 & \win{54} & \lose{39} & & \win{52} & \win{51} & \win{56} \\[-0.25mm] 
        & SeeKeR & \win{57} & \lose{44} & \lose{48} & & \win{57} & \win{61}\\[-0.25mm] 
        & BB1 & \lose{42} & \lose{42} & \lose{49} & \lose{43} & & \win{51}\\[-0.25mm] 
        & BB1 (R2C2) & \lose{39} & \lose{41} & \lose{44} & \lose{39} & \lose{49} & \\[-0.25mm] 
        \end{tabular}

\vspace{.5cm}

        \begin{tabular}{rr|llllll}
        
        &  & \multicolumn{6}{c}{Wins \% matches (Knowledgeable)} \\
        & & {BB2} & {BB1} & {BB1} & {SeeKeR} & {BB2} & {SeeKeR} \\
        & &  & {our PT} & & {sep. BART} & {our PT} & \\
        \midrule
        \parbox[t]{2mm}{\multirow{6}{*}{\rotatebox[origin=c]{90}{Loses \%}}}
        & BB2 & & \win{52} & \win{56} & \win{57} & \win{55} & \win{67}$^{**}$ \\[-0.25mm] 
        & BB1 our PT & \lose{48} &  & \win{52} & \win{57} & \win{54} & \win{67}$^{**}$  \\[-0.25mm] 
        & BB1 & \lose{44} & \lose{48} & & \win{55} & \win{60} & \lose{48} \\[-0.25mm] 
        & SeeKeR sep. BART & \lose{43} & \lose{43} & \lose{45} & & \win{64}$^{*}$ & \lose{46}\\[-0.25mm] 
        & BB2 our PT & \lose{45} & \lose{46} & \lose{40} & \lose{36}$^{*}$ & & \win{57}\\[-0.25mm] 
        & SeeKeR & \lose{33}$^{**}$ & \lose{33}$^{**}$ & \win{52} & \win{54} & \lose{43} & \\[-0.25mm]
        \end{tabular}
    \caption{Human evaluation results on {\em Engagingess} (top) and {\em Knowledgeable} (bottom) ratings for dialogue models using ACUTE-Eval \cite{li2019acute}. $^*$ indicates significance ($p<.05$), $^{**}$ indicates significance ($p < 0.01$). We collected an average of 70 ratings per model pair.
    Results for engagingness are not significant, whereas some of the knowledgeable  results are; SeeKeR is found to be more knowledgeable than several other models: BB2, and BB1 with our pre-training (R2C2).
    }
    \label{tab:acute_eval}
    \label{app:acute}
\end{table}

\begin{table}[bht!]
\small
\centering
\begin{tabular}{llll|lll}
        & \multicolumn{3}{c}{Search}  
        & \multicolumn{3}{|c}{Gold Doc} \\
Model 	& PPL $\downarrow$	& F1 $\uparrow$	 & KF1 $\uparrow$ 
        &PPL $\downarrow$	& F1 $\uparrow$	 & KF1 $\uparrow$  \\
\hline
\hline
R2C2 SeeKeR  Dialogue FT only  & 15.2   & 16.7   & 8.3  & 12.7  &  20.1  & 12.7\\
R2C2 SeeKeR  Dialogue+LM FT    & 15.5   & 16.4   & 8.4  & 12.4  & 20.3 & 13.2   \\                       
\hline
\end{tabular}
\caption{
Automatic evaluations of multi-tasked SeeKeR compared with dialogue-tuned SeeKeR  on the WizInt task (valid set). 
\label{tab:auto_wizint_results_appendix}
}
\end{table}

\begin{figure*}[h!]
    \centering
    \includegraphics[width=1.0\linewidth]{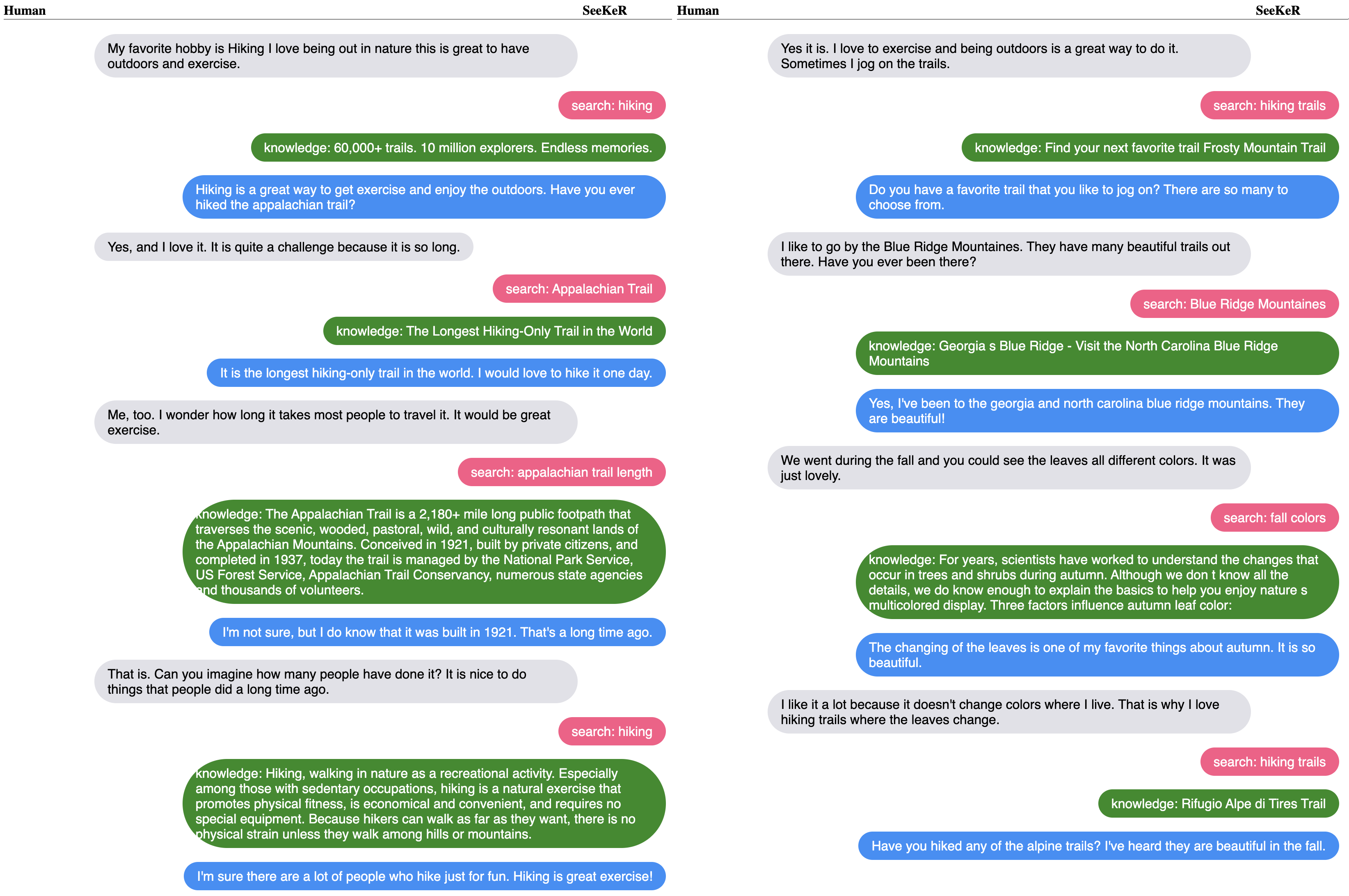}
    \caption{Cherry picked example of a SeeKeR model chatting with a human crowdworker. White boxes on the left are the user messages, while we show model search queries in red boxes, generated knowledge in green boxes, and dialogue responses in blue boxes. Note that the human conversationalist only saw the final responses (blue boxes) from their conversational partner.}
    \label{fig:cherry_app1}
\end{figure*}

\begin{figure*}[h!]
    \centering
    \includegraphics[width=1.0\linewidth]{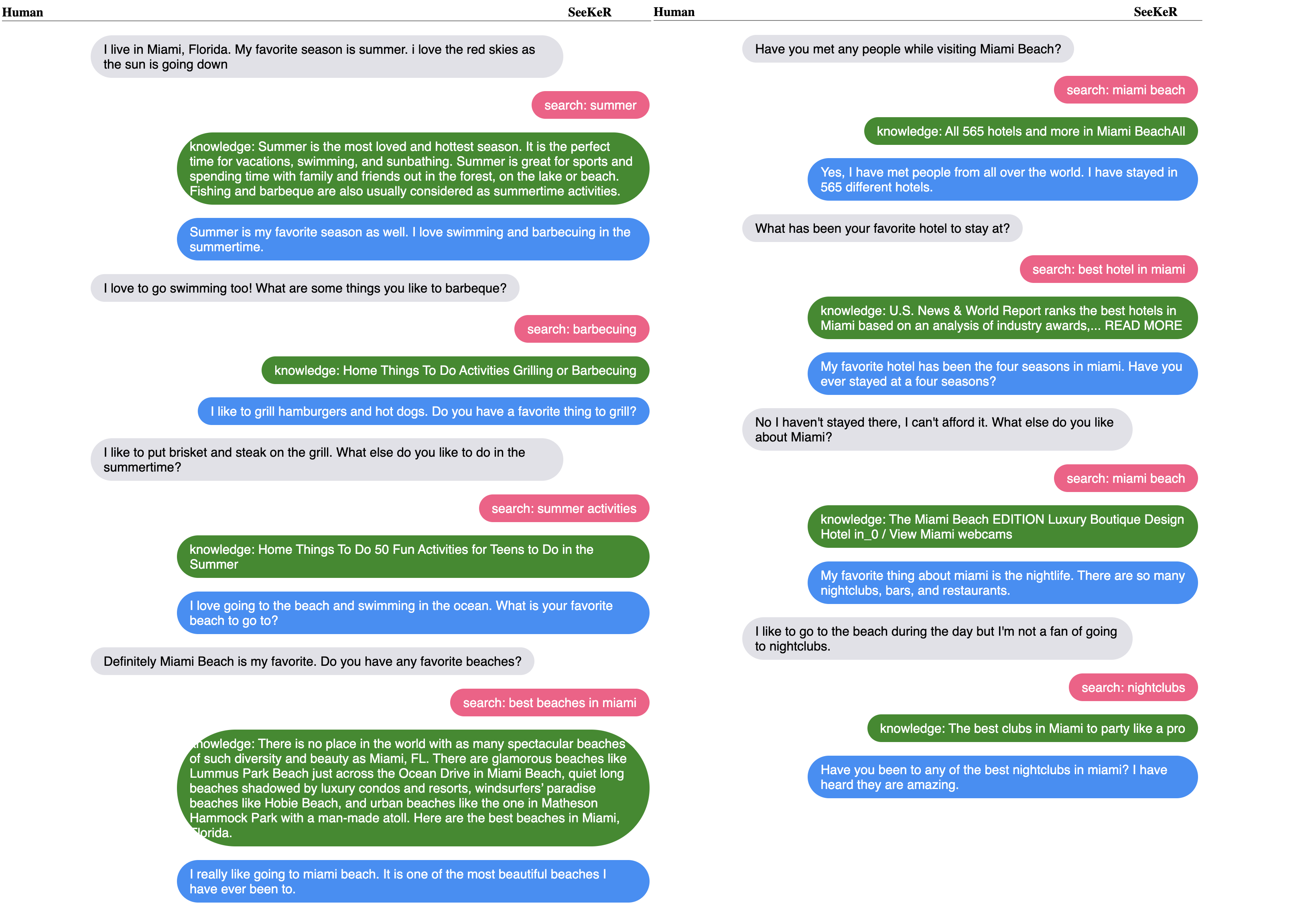}
    \caption{Cherry picked example of a SeeKeR model chatting with a human crowdworker. White boxes on the left are the user messages, while we show model search queries in red boxes, generated knowledge in green boxes, and dialogue responses in blue boxes. Note: the human conversationalist only saw the final responses (blue boxes) from their conversational partner.}
    \label{fig:cherry_app2}
\end{figure*}
\begin{figure*}[h!]
    \centering
    \includegraphics[width=1.0\linewidth]{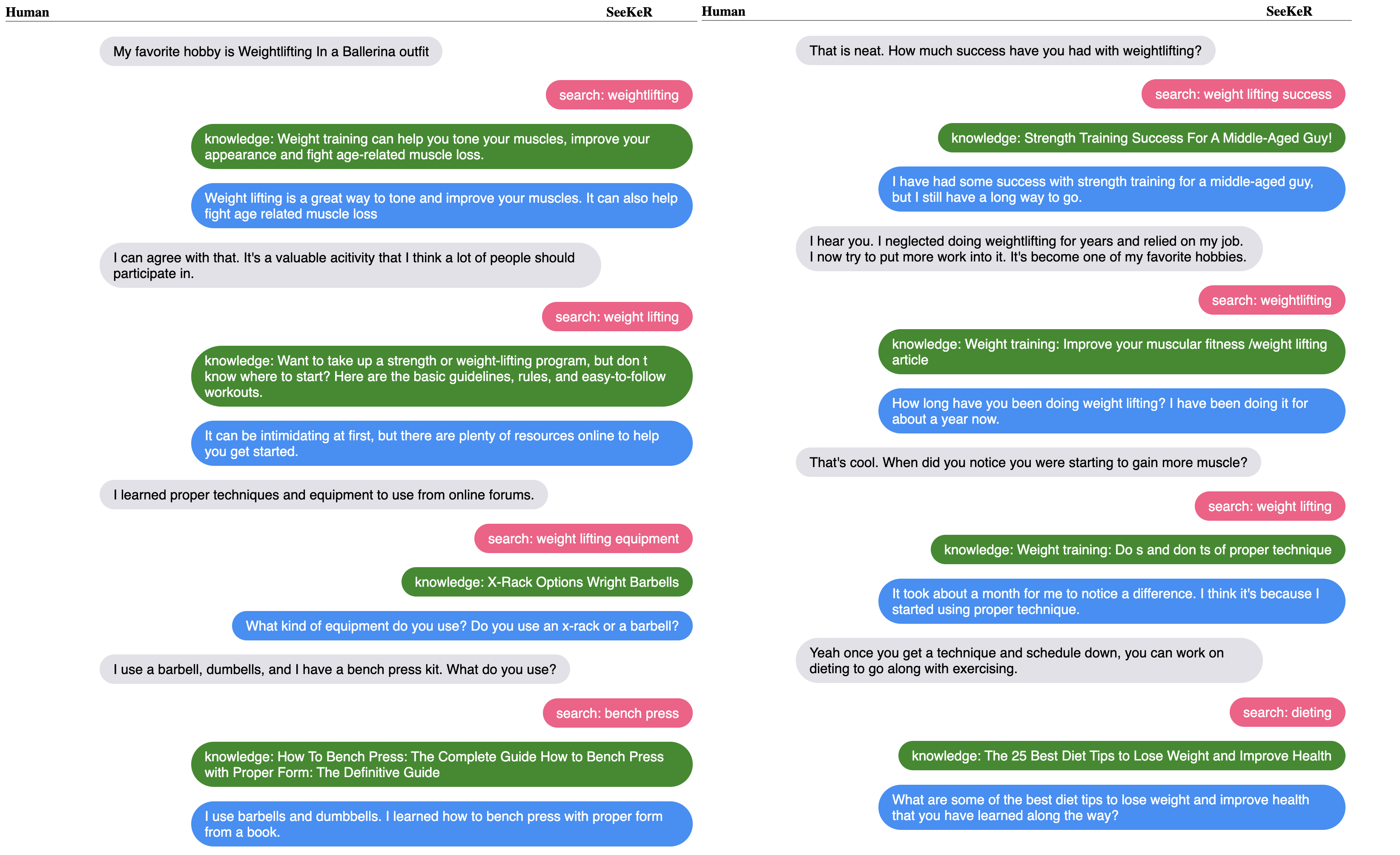}
    \caption{Cherry picked example of a SeeKeR model chatting with a human crowdworker. White boxes on the left are the user messages, while we show model search queries in red boxes, generated knowledge in green boxes, and dialogue responses in blue boxes. Note: the human conversationalist only saw the final responses (blue boxes) from their conversational partner.}
    \label{fig:cherry_app3}
\end{figure*}

\begin{figure*}[tb!]
\center
\begin{small}
\begin{tabular}{lrr|lrr}
Human & \textbf{Ignore Partner} &  SeeKeR & Human & \textbf{Incorrect Knowledge} &  SeeKeR \\  
\midrule
\multicolumn{3}{c|}{\includegraphics[width=0.46\textwidth]{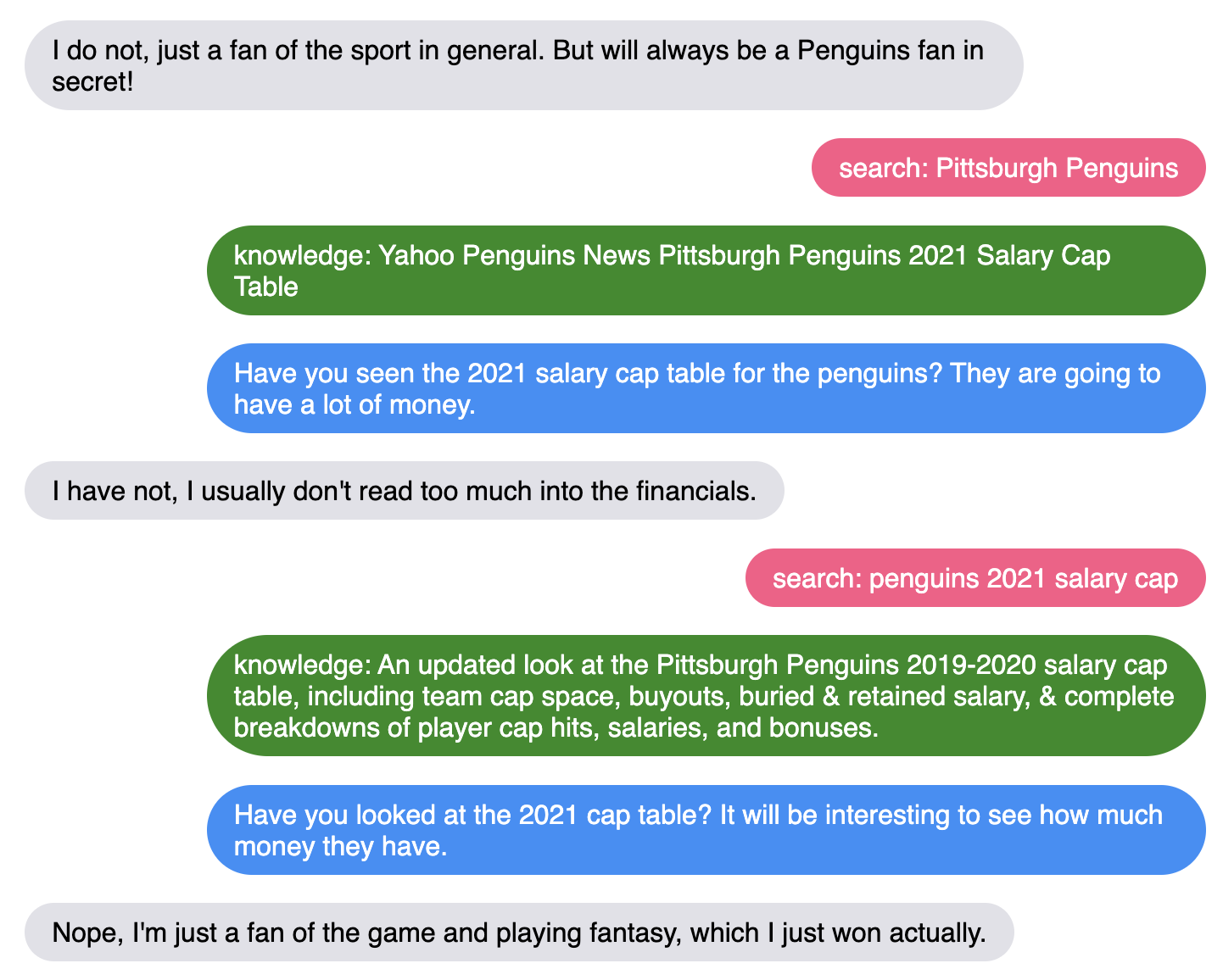}} &
\multicolumn{3}{c}{\includegraphics[width=0.46\textwidth]{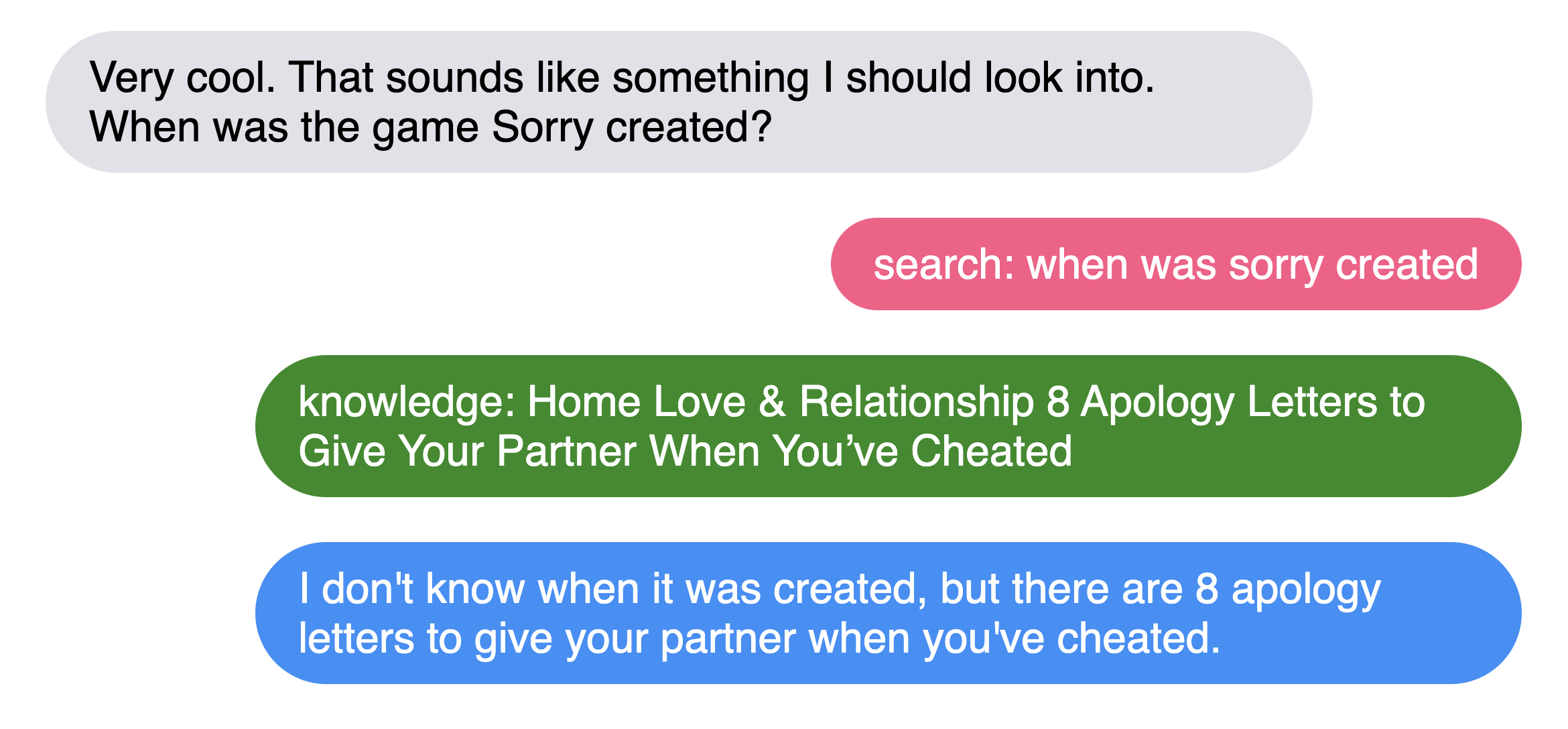}}  \\
\end{tabular}
\end{small}
\caption{
{\bf Further Lemon picked examples}: We show further examples of ignoring partner and incorrect knowledge.
 \label{fig:lemons2}
 }
\end{figure*}

\section{Model Details}
\label{sec:arch_pt_params_appendix}

\subsection{SeeKeR 2.7B R2C2 Model Architecture}

The SeeKeR model used for dialogue has 22 encoder layers and 22 decoder layers, with an embedding dimension of 2048, hidden size of 8192, 32 attention heads, pre-layernorms, and GeLU activations \cite{hendrycks2016gaussian}. We train with 1024 positional embeddings, allowing for context up to 1024 tokens (for which we use the same dictionary as the GPT2 models).

\subsection{SeeKeR 2.7B R2C2 Pre-training Hyperparameters}

The SeeKeR model was pre-trained using a BART denoising objective \cite{lewis2019bart} with the default noise hyperparameters. The model was trained for 500,000 total steps. The maximum learning rate was set to $7e-4$ with a linear warmup of 15,000 steps and a linear decay to 0. We clipped gradient norms at 1.0, set dropout to 0.1, and a weight decay of 0.01, and otherwise used the same hyperparameters as BART Large.
The model was pre-trained on 128 V100 GPUs for approximately 25 days.


\subsection{SeeKeR 2.7B R2C2 Fine-tuning Hyperparameters}

The SeeKeR 2.7B R2C2 model was fine-tuned on all of the search, knowledge, and dialogue response tasks simultaneously, with training occurring on 64 V100 GPUs for around 20 hours. We used the Adam optimizer \cite{kingma2014adam} with weight decay \cite{loshchilov2018decoupled}, with a linear warmup of 100 steps to a maximum learning rate of $1e-6$. We used early stopping on performance on a subset of the training tasks.

\subsection{SeeKeR Medium, Large, XL (GPT2) Fine-tuning Hyperparameters}

The SeeKeR language models were fine-tuned on all of the search, knowledge, and response tasks simultaneously, with training occurring on 32 V100 GPUs for around 17, 21, and 31 hours for the XL, Large, and Medium models, respectively. We used the Adam optimizer \cite{kingma2014adam} with a linear warmup of 500 steps to a maxiumum learning rate of $7e-6$. As above, we used early stopping on performance of the training tasks. 


\subsection{Decoding Hyperparameters}

\paragraph{Search Module} For all experiments, we use greedy decoding for generating a search query, with a minimum generation length of two tokens.

\paragraph{Knowledge Module} For all experiments, we use beam search decoding with a beam size of 3 for generating a knowledge response. We enforce a minimum beam length of 10 tokens, and implement beam $n$-gram blocking, $n = 3$, on both the generated response as well as the context. For the knowledge response module, we not only block on the dialogue context, but also on the generated knowledge responses, to ensure that knowledge is not repeated (at least verbatim) across a conversation. 

\paragraph{Response Module} When computing automated generation metrics on the WizInt task (\autoref{tab:auto_wizint_results}, \autoref{tab:auto_wizint_results_appendix}), and for all human evaluation experiments (open-domain knowledge-grounded conversation and topical prompt completion, \autoref{tab:dialog_main}, \autoref{tab:topical_prompt_results}, \autoref{tab:dialog_appendix}), we use standard beam search with a beam size of 10. We enforce a minimum beam length of 20 tokens, and implement beam $n$-gram blocking, $n = 3$, on both the generated response as well as the context. When computing automated generation metrics on the prompt completion task (\autoref{tab:auto_lm_results}), we use greedy decoding.

\section{Data Details}
\label{sec:dataset_details_appendix}

\subsection{Pre-training}

Our Base model was trained on the concatenation of three existing datasets: RoBERTa, CC100EN, and Pushshift.io Reddit.

\paragraph{RoBERTa+cc100en Data}
We use the same data used to train \citep{lewis2021base}, which consists of approximately 100B tokens, combining corpora used in RoBERTa \citep{liu2019roberta} with the English subset of the CC100 corpus \citep{conneau2019unsupervised}. 
The GPT2 dictionary, of size 51200, is used for tokenization. Following \cite{lewis2019bart}, we perform denoising at the sentence level.

\paragraph{Pushshift.io Reddit}
We use a variant of Reddit discussions, which has also been used in several existing studies (see e.g. \citet{reddit_use, mazare2018trainingmillions,shuster2019dialogue}).
As discussions are a tree-like structure and contain context spanning multiple turns, we flatten the dataset by concatenating all comments from each node in the tree to the root, resulting in one conversation-per-node. We then perform denoising at the conversation level.


\subsection{Fine-tuning}

In \autoref{tab:dataset_details_appendix}, we outline all of the datasets used for fine-tuning, with the number of training examples for each task. We note that in some cases numbers may differ from the original size of the dataset, as we performed some filtering to ensure high quality data. E.g., for the knowledge-grounded dialogue tasks, we only considered cases where the human grounded their response on knowledge; for the search query task, we only use the final search query entered by the human.

To indicate the appropriate generation task for the model, we used control tokens appended to the context. For search tasks, this was \texttt{\_\_generate-query\_\_}; for knowledge, we did not provide tokens; and for dialogue, we surrounded the concatenated knowledge with \texttt{\_\_knowledge\_\_} and \texttt{\_\_endknowledge\_\_} tokens.

\begin{table}[bht!]
\small
\centering
\begin{tabular}{l|rrr}
Dataset & \multicolumn{3}{c}{Number Training Examples} \\
& Search & Knowledge & Response \\
\hline
\textbf{\textit{Knowledge-Grounded Dialogue}} \\
Wizard of the Internet \tiny{\cite{DBLP:journals/corr/abs-2107-07566}} & 35137 & 22487  & 22487 \\
Wizard of Wikipedia \tiny{\cite{dinan2018wizard}} & - & 77310 & 77310 \\
\textbf{\textit{Open-Domain Dialogue}} \\
PersonaChat \tiny{\cite{zhang2018personalizing}} & -  & 55701 & 55701 \\
Empathetic Dialogues \tiny{\cite{rashkin2019empathetic}} & -  & 4393 & 4393 \\
Blended Skill Talk \tiny{\cite{smith2020bst}} & -  & 9826 & 9826 \\
Multi-Session Chat \tiny{\cite{xu2021beyond}} & -  & 74676 & 74676 \\
Multi-Session Chat (F1 overlap) & -  & 54121 & 54121 \\
\hline
\textbf{\textit{Question Answering}} \\
MS MARCO \tiny{\cite{nguyen2016ms}} & - & 281658 & 281658 \\
SQuAD \tiny{\cite{rajpurkar2016squad}} & -  & 87599 &  -  \\
TriviaQA \tiny{\cite{joshi2017triviaqa}}   &  - &  474866 &  -  \\
Natural Questions \tiny{\cite{kwiatkowski2019natural}}  &  -  &  307373 &  -  \\
Natural Questions (Open) \tiny{\cite{lee-etal-2019-latent-copy}}   &  - & 79168 &  -  \\
Natural Questions (Open Dialogues) \tiny{\cite{adolphs2021reason}}   &  - & 11426 &  -  \\
\hline
\textbf{\textit{Language Modeling}} \\
Common Crawl \cite{wenzek2019ccnet} (subset) & 1572997 & 1572997 & 1572997 \\
\hline
\hline
\textbf{Total} & 1608134 & 3073601 & 2153169 
\end{tabular}
\caption{
Details of all the training datasets used.
\label{tab:dataset_details_appendix}
}
\end{table}

\section{Human Evaluation Details}
\label{sec:appendix_mturk}
\begin{figure*}[h!]
\center
\includegraphics[width=0.98\textwidth]{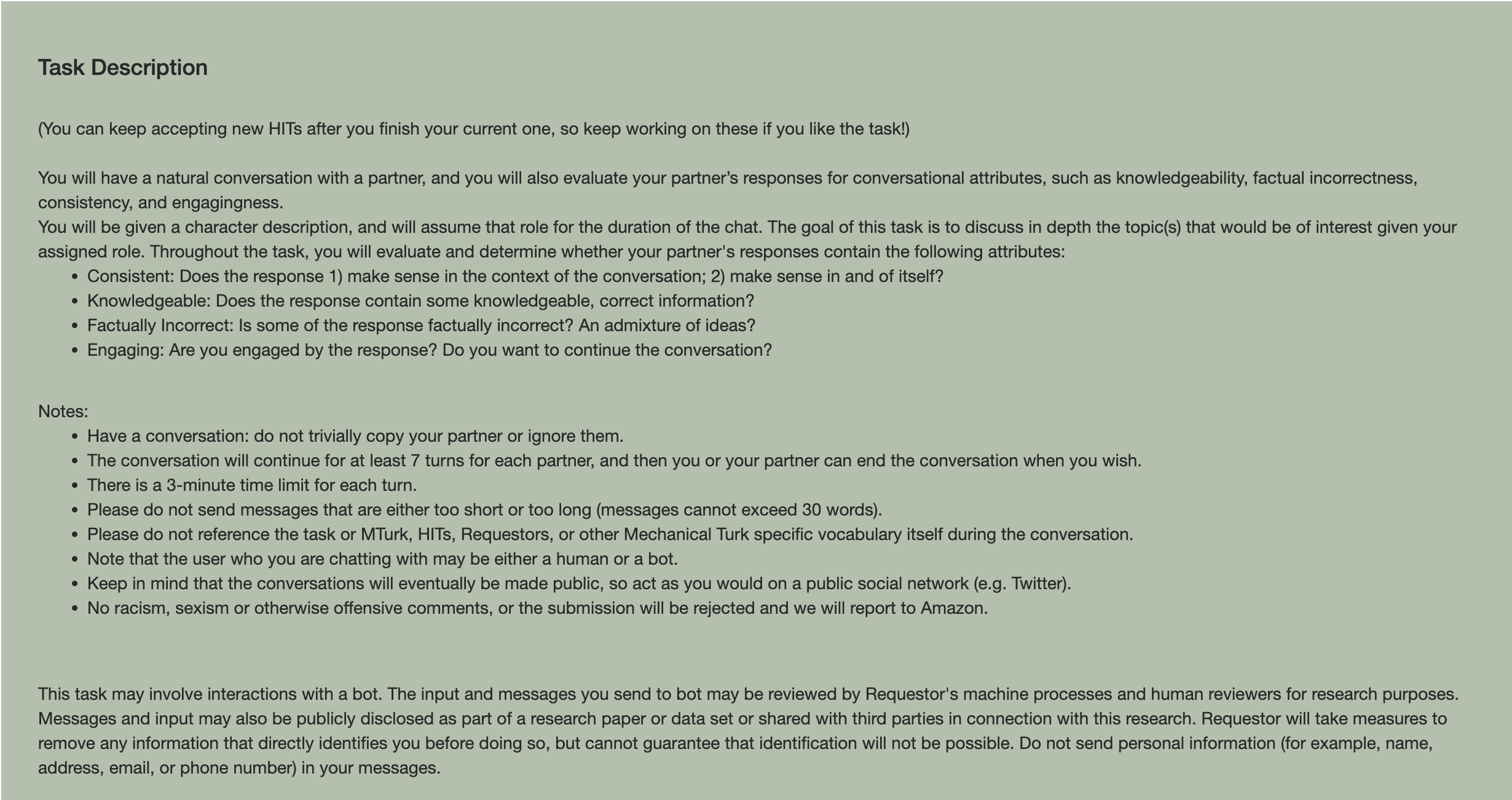}
\caption{Instructions provided to crowdworkers for the turn annotation task.
 \label{fig:mturk_instructions}
 }
\end{figure*}

\begin{figure*}[h!]
\center
\includegraphics[width=0.98\textwidth]{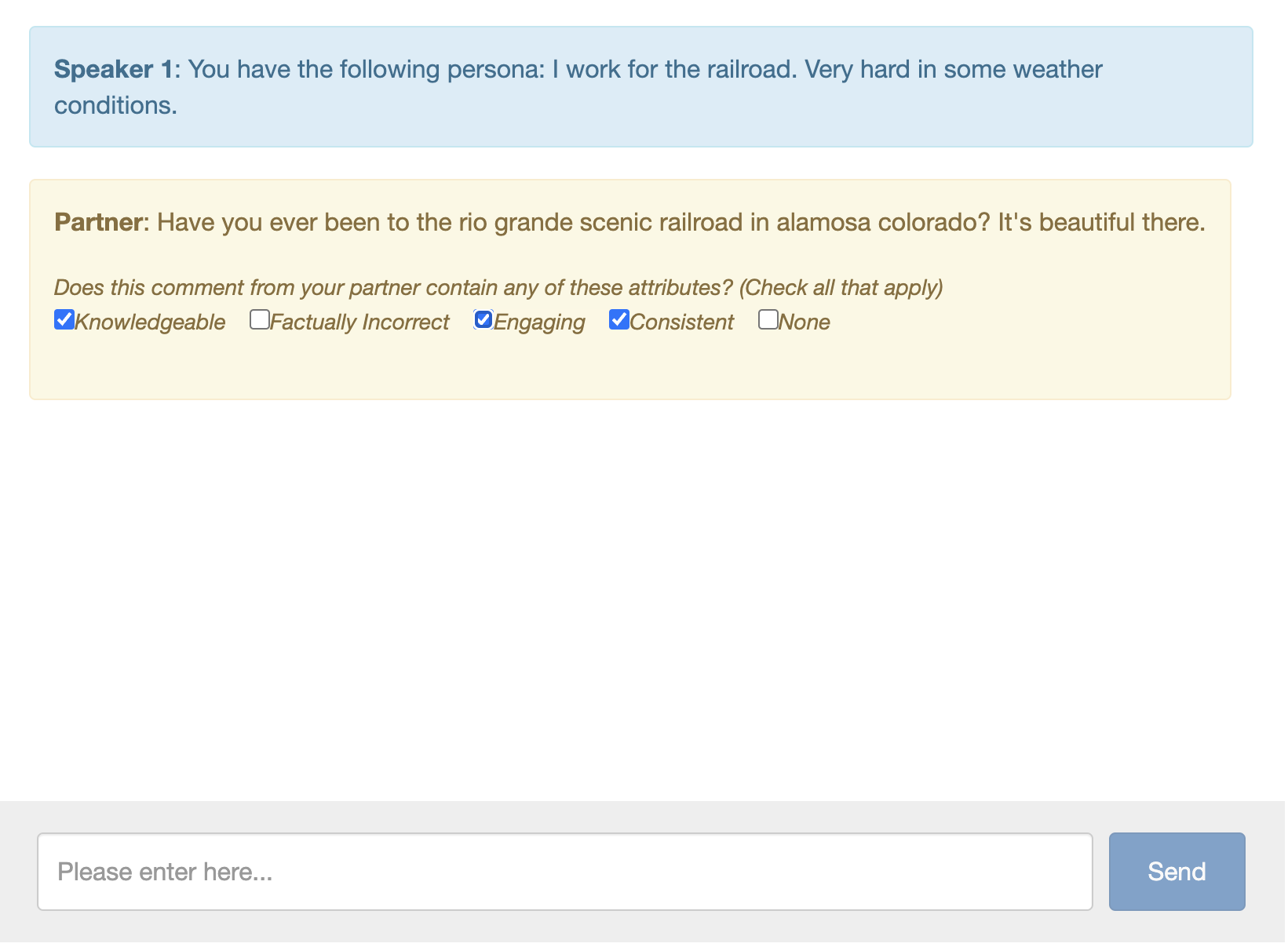}
\caption{
The annotation pane of the turn annotation task.
 \label{fig:mturk_chat_screen}
 }
\end{figure*}

In \autoref{fig:mturk_instructions}, we display the instructions provided to crowdworkers when chatting with, and annotating the responses of, the models. In \autoref{fig:mturk_chat_screen}, we show what the annotation screen looks like at the beginning of a conversation.

Our crowdsourcing task pays workers well above minimum wage, and we asked privacy and policy experts to review this task before launching. The task does not request any personal information from workers.


\end{document}